\documentclass[lettersize,journal]{IEEEtran}

\usepackage{graphicx}
\usepackage{subcaption}
\usepackage{amsmath}
\usepackage{algorithm}
\usepackage{algpseudocode}
\usepackage{xcolor}
\usepackage{amssymb}

\usepackage{graphicx}

\ifCLASSOPTIONcompsoc
  \usepackage[nocompress]{cite}
\else
  \usepackage{cite}
\fi
\begin{document}
\title{Decentralised Resource Sharing in TinyML: Wireless Bilayer Gossip Parallel SGD for Collaborative Learning}

\author{Ziyuan Bao, 
        Eiman Kanjo,
        Soumya Banerjee,
        Hasib-Al Rashid,
        and~Tinoosh Mohsenin
        
\thanks{Z. Bao and E. Kanjo are with the Department of Computing, Imperial College London, United Kingdom. E. Kanjo is also with Nottingham Trent University, United Kingdom. (email: z.bao@imperial.ac.uk, eiman.kanjo@ntu.ac.uk, eiman.kanjo@ntu.ac.uk).}

\thanks{S. Banerjee is with Trasna-Solutions Ltd., Rousset, France (email: soumya.banerjee@trasna.io).}

\thanks{H. Rashid is with the Department of Computer Science and Electrical Engineering, University of Maryland Baltimore County, Baltimore, MD, USA (email: hrashid1@umbc.edu).}

\thanks{T. Mohsenin is with the Department of Electrical and Computer Engineering, Johns Hopkins University, Baltimore, MD, USA (email: tinoosh@jhu.edu).}

}




\maketitle

\begin{abstract}
With the growing computational capabilities of microcontroller units (MCUs), edge devices can now support machine learning models. However, deploying decentralised federated learning (DFL) on such devices presents key challenges, including intermittent connectivity, limited communication range, and dynamic network topologies. This paper proposes a novel framework, bilayer Gossip Decentralised Parallel Stochastic Gradient Descent (GD-PSGD), designed to address these issues in resource-constrained environments. The framework incorporates a hierarchical communication structure using Distributed K-means (DK-means) clustering for geographic grouping and a gossip protocol for efficient model aggregation across two layers: intra-cluster and inter-cluster. We evaluate the framework’s performance against the Centralised Federated Learning (CFL) baseline using the MCUNet model on the CIFAR-10 dataset under IID and Non-IID conditions. Results demonstrate that the proposed method achieves comparable accuracy to CFL on IID datasets, requiring only 1.8 additional rounds for convergence. On Non-IID datasets, the accuracy loss remains under 8\% for moderate data imbalance. These findings highlight the framework’s potential to support scalable and privacy-preserving learning on edge devices with minimal performance trade-offs.

\end{abstract}


%
\IEEEpeerreviewmaketitle

\section{Introduction}
The rapid evolution of machine learning and AI has substantially increased the volume of data generated by edge devices, such as mobile phones, IoT sensors, and wearable devices. This data explosion has highlighted the limitations of traditional centralised learning models, where data is aggregated at a central server for model training. These centralised models face significant challenges, including concerns about data privacy, high bandwidth consumption, and increased latency, making them unsuitable for many edge-based applications~\cite{wen2023survey}.
To address these issues, Federated Learning (FL) was introduced by McMahan et al.~\cite{mcmahan2017fl}. FL enables edge devices to collaboratively train a global model while retaining their local data, thereby preserving privacy and reducing communication costs. However, most FL implementations rely on a Centralised Federated Learning (CFL) paradigm, where a central server is responsible for model aggregation. This centralised approach creates risks, such as bottlenecks in scalability, increased latency, and a single point of failure, which undermines the robustness of the system~\cite{li2020fl}.
Decentralised Federated Learning (DFL) has emerged as an alternative to CFL. By enabling direct communication between edge devices, DFL eliminates the need for a central server, improving scalability and resilience~\cite{yuan2024dfl}. Nevertheless, existing DFL frameworks often assume static and fully connected network topologies, which are impractical in real-world scenarios. 
In traditional federated learning, the central server often waits for all participating devices to finish their computations and submit updated weights. This results in idle time for the server, creating unnecessary delays in the training process. The situation becomes critical if the central server, which coordinates cumulative learning and aggregates model-updates fails. Such a failure disrupts the entire learning workflow, causing system downtime and halting progress.
Furthermore, edge devices exhibit practical constraints, including limited communication ranges, low computational capabilities, and dynamic network conditions. These challenges severely limit the feasibility and effectiveness of current decentralised federated learning (DFL) methods in real-world applications~\cite{lu2024noniid,wang2019adaptive}.
To overcome these limitations, we propose a bilayer Gossip Decentralised Parallel Stochastic Gradient Descent (GD-PSGD) framework. This framework introduces a hierarchical communication structure that combines geographic clustering and gossip-based model aggregation. Distributed K-means (DK-means) for efficient clustering and leveraging gossip protocols for intra- and inter-cluster communication help in reducing communication overhead and improving scalability in resource-constrained environments.
The key contributions of this paper are as follows:
\begin{itemize}
    \item A bilayer gossip protocol that uses Distributed K-means (DK-means) for efficient geographic clustering.
    \item A decentralised communication model tailored to edge environments, addressing limited communication ranges and dynamic topologies.
    \item Performance and Comprehensive evaluation of the Framework   in IID and Non-IID data scenarios, demonstrating competitive results compared to CFL.
\end{itemize}

\section{Related Work}
Federated Learning (FL) has gained significant attention as a solution for privacy-preserving distributed machine learning. Originally proposed by Google researchers, FL allows multiple devices to collaboratively train a shared global model while keeping their local data private~\cite{mcmahan2017fl}. Early implementations of FL demonstrated its potential in applications such as personalised language models for mobile devices, including Google's keyboard, where local data remained secure on user devices~\cite{wen2023survey}.
Traditional FL architectures are typically centralised, with a server coordinating model-updates and aggregation~\cite{mcmahan2017fl}. Methods such as Federated Averaging (FedAvg) have been widely adopted for global model updates, ensuring efficiency and scalability~\cite{li2020fl}. However, centralised federated learning (CFL) faces challenges such as single points of failure, where server outages disrupt the entire learning cycle, and communication bottlenecks, as the central server's communication load increases with the number of devices, limiting scalability~\cite{yuan2024dfl}. These limitations have motivated the development of decentralised federated learning (DFL) as a more robust alternative~\cite{anusha_dfl}.
DFL eliminates the reliance on a central server by enabling peer-to-peer communication for model-updates. This decentralised approach improves scalability and system resilience~\cite{yuan2024dfl}. For instance, frameworks based on peer-to-peer communication significantly reduce communication overhead and enhance scalability by eliminating the central server~\cite{lian2017dpsgd}. However, the absence of a central coordinator introduces new challenges, including the need for efficient network topologies and communication protocols~\cite{matcha}. Fully connected mesh topologies ensure robustness but incur high communication costs, making them impractical for edge devices with limited resources~\cite{lu2024noniid}. Simpler topologies like rings or stars are less communication-intensive but can result in delays and inefficiencies in sparse or dynamic networks~\cite{wang2019adaptive}.
Adaptive topologies have been explored to address these challenges, often leveraging clustering techniques to group devices based on proximity or data similarity. Hierarchical frameworks combine the benefits of clustering and decentralisation, facilitating efficient communication and aggregation~\cite{mcunet}. These designs are further enhanced by gossip-based protocols, which achieve strong convergence rates by iteratively aggregating model updates across peers. For instance, the MATCHA algorithm dynamically schedules non-overlapping communication links to reduce overhead while maintaining connectivity~\cite{matcha}. Such methods demonstrate the potential of decentralised architectures for handling large-scale deployments~\cite{mcmahan2017fl}.
Recent work has also focused on enabling FL in resource-constrained environments. On-device training and TinyML techniques, such as model quantisation and pruning, allow FL models to run on microcontrollers and other low-power edge devices. MCUNet exemplifies this direction by enabling efficient model deployment on resource-limited hardware, paving the way for scalable FL in IoT applications~\cite{mcunet}.
Decentralised optimisation methods, such as Decentralised Parallel Stochastic Gradient Descent (D-PSGD), further enhance DFL by removing the dependency on a central server~\cite{lian2017dpsgd}. These methods ensure stable convergence even under dynamic network conditions by leveraging gradient tracking and variance reduction techniques. Frameworks incorporating D-PSGD demonstrate performance comparable to centralised methods, making them a viable alternative for large-scale, decentralised learning systems~\cite{li2020fl}.
The integration of advanced communication protocols, adaptive topologies, and lightweight models underscores the evolving landscape of FL. These advancements aim to address the inherent challenges of decentralisation, such as network dynamism, resource constraints, and data heterogeneity, while maintaining robust model performance.

\begin{figure}[h!]
    \centering
    \begin{subfigure}[t]{0.48\linewidth}
        \centering
        \includegraphics[width=\linewidth]{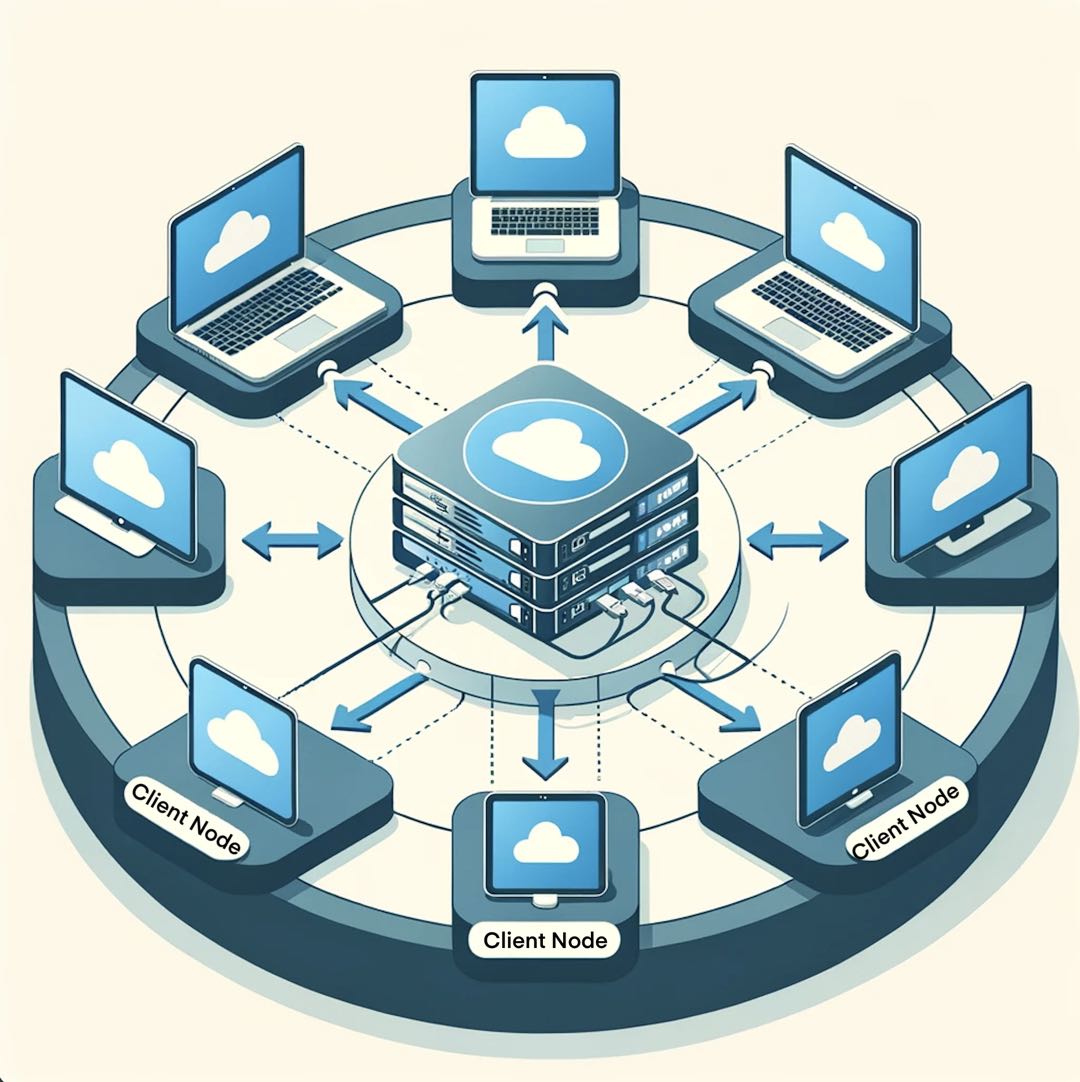}
        \caption{Centralised Federated Learning Architecture.}
        \label{fig:cfl}
    \end{subfigure}
    \hfill
    \begin{subfigure}[t]{0.48\linewidth}
        \centering
        \includegraphics[width=\linewidth]{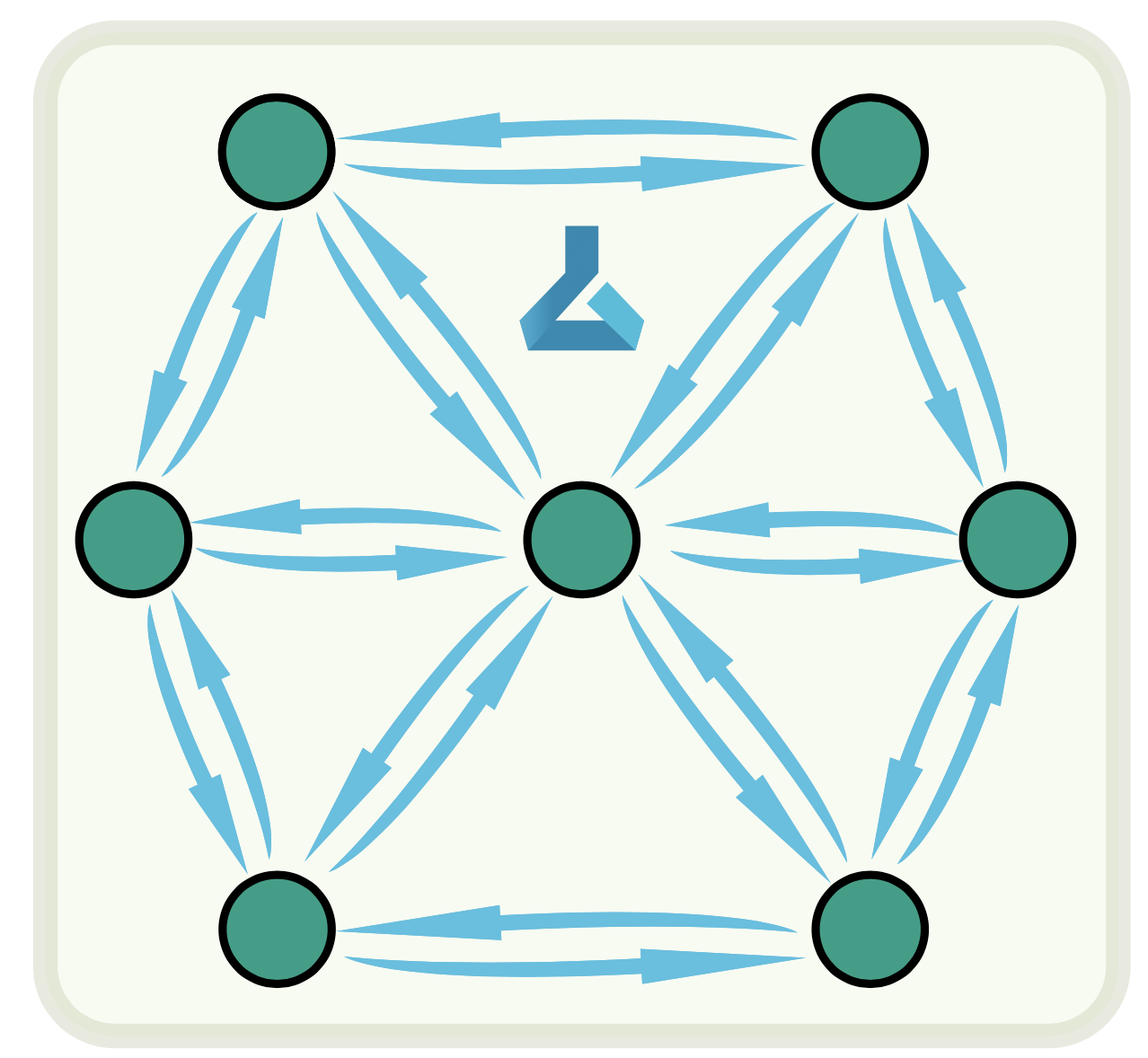}
        \caption{The DFL architecture.}
        \label{fig:DFL}
    \end{subfigure}
    \caption{Comparison of Centralised Federated Learning and Decentralised Federated Learning Architectures.}
    \label{fig:comparison}
\end{figure}

There are several topology options applicable to Decentralised Federated Learning (DFL), including line, ring, mesh (fully connected), star, and hybrid. The choice of topology greatly impacts information transmission and model aggregation and must consider the deployment environment and the capabilities of the devices involved.
In a linear topology, all client devices are connected in a fixed order, with each device having a specific predecessor and successor. Information propagation occurs sequentially: device 1 updates its model and passes it to device 2, which aggregates and trains a new model before passing it to device 3, and so on. While simple and easy to implement, this topology is highly vulnerable to single points of failure and cannot perform parallel updates, resulting in inefficiency~\cite{yuan2024dfl}. Similarly, a ring topology works like a linear topology but loops indefinitely, allowing models to continuously adapt to new data until convergence.
\vspace{0.5em}
Mesh topology, in contrast, connects all nodes to each other, enabling updates to be broadcasted to all other nodes. This structure eliminates single points of failure but imposes significant strain on communication bandwidth. Each device must send its parameters to all others during each training round, resulting in communication overhead proportional to the total number of devices. This creates challenges for edge devices, which often have limited bandwidth and computational resources. Despite these limitations, approximately half of DFL-related research assumes a fully connected network due to its simplicity~\cite{lian2017dpsgd,matcha}. However, practical deployment often requires alternatives due to the limited communication bandwidth and range of edge devices~\cite{mcunet}.
\vspace{0.5em}
Star topology resembles the structure of Centralised Federated Learning (CFL), with a central client acting as the server. Unlike a central server in CFL, the central client can train its own local model and collect local data. While this approach reduces complexity, it still faces bandwidth limitations and cannot scale well in environments with many devices~\cite{10.1145/3659205}.
\vspace{0.5em}
Hybrid topology offers greater flexibility by combining elements from different topologies. For example, devices can be divided into clusters where a ring topology is used within clusters, and a leader node from each cluster connects to other cluster leaders using a star or mesh topology. Clustering algorithms may group devices based on characteristics such as data similarity, geographic proximity, or computing power. Hybrid structures optimise bandwidth usage while maintaining scalability by limiting cluster sizes, allowing for effective communication even in constrained environments~\cite{mcmahan2017fl,wen2023survey}.
\vspace{0.5em}
On-device training addresses the computational limitations of edge devices while balancing privacy and performance. Cloud-based methods pose risks to data privacy and increase latency due to network dependency, while inference-only methods restrict model adaptability. On-device training solves these issues by allowing local model updates, ensuring that data remains private and eliminating transmission delays. Edge devices can quickly adapt models to new data, improving customisation and real-time responsiveness~\cite{alajlan2022tinyml}. 
TinyML refers to the field of machine learning technologies and applications capable of performing on-device analytics on resource-constrained devices. The development of TinyML is driven by several key factors, including advancements in hardware, optimized algorithms, and supportive software ecosystems. Modern microcontrollers have become more powerful and efficient, allowing for the execution of lightweight machine learning models \cite{lê2023efficientneuralnetworkstiny} ~\cite{alajlan2022tinyml}. Concurrently, techniques such as quantization, pruning, and knowledge distillation have emerged to reduce the size and computational requirements of machine learning models without significantly compromising their accuracy. Additionally, the development of neural network architectures specifically tailored for edge devices, such as MobileNets and MCUNet, ensures that complex tasks can be performed within the constraints of limited hardware resources \cite{lin2020mcunet} without compromising model accuracy~\cite{shi2022decentralized}..
\begin{figure}[h!]
    \centering
    \begin{subfigure}[t]{0.22\linewidth}
        \centering
        \includegraphics[width=\linewidth]{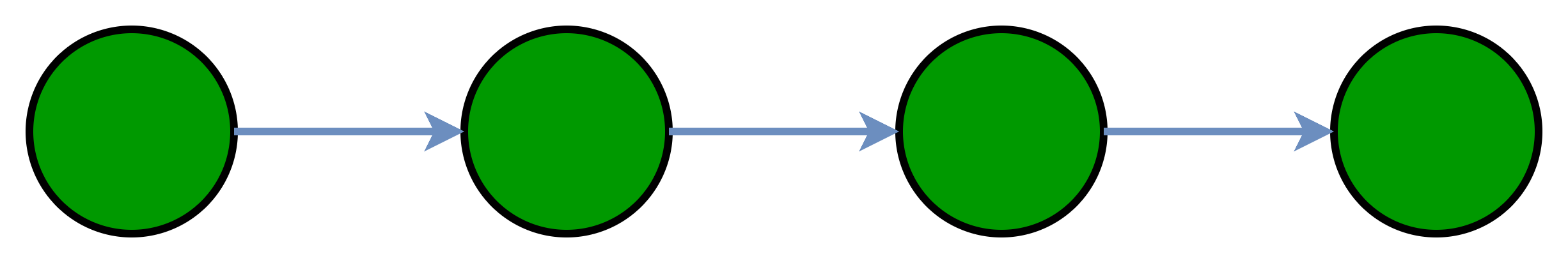}
        \caption{Line}
        \label{fig:line_topo}
    \end{subfigure}
    \hfill
    \begin{subfigure}[t]{0.22\linewidth}
        \centering
        \includegraphics[width=\linewidth]{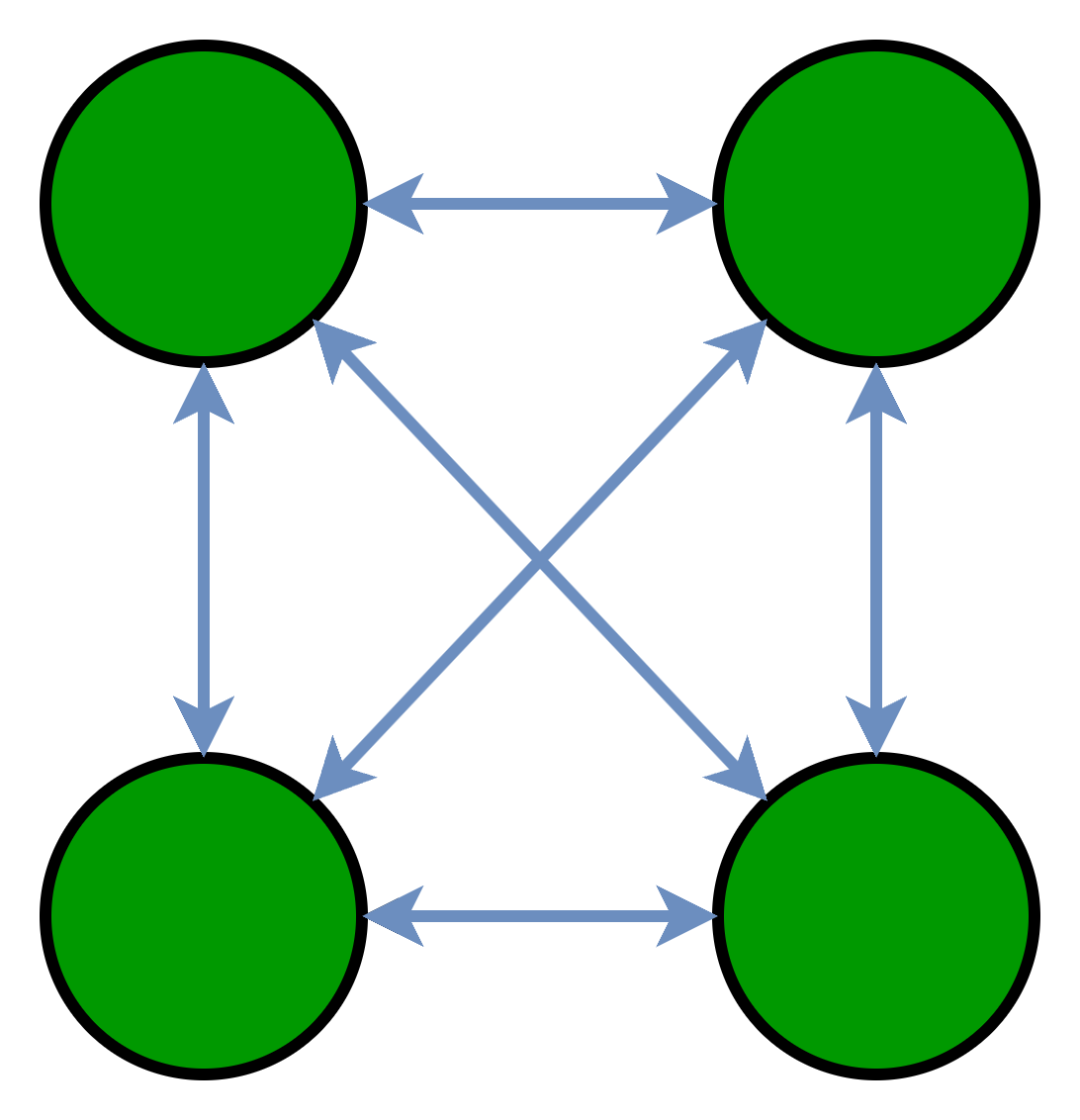}
        \caption{Mesh}
        \label{fig:mesh_topo}
    \end{subfigure}
    \hfill
    \begin{subfigure}[t]{0.22\linewidth}
        \centering
        \includegraphics[width=\linewidth]{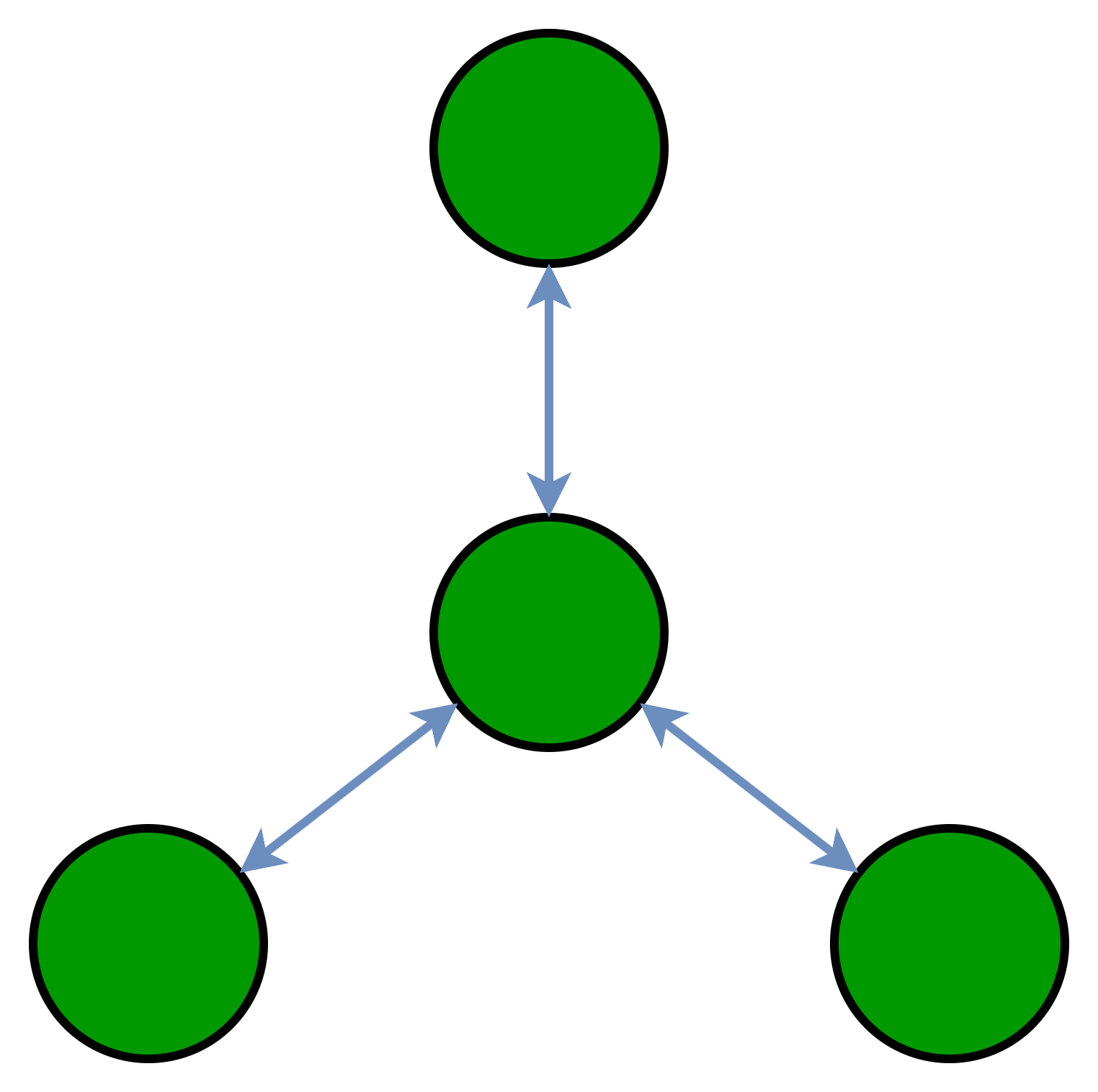}
        \caption{Star}
        \label{fig:star_topo}
    \end{subfigure}
    \hfill
    \begin{subfigure}[t]{0.22\linewidth}
        \centering
        \includegraphics[width=\linewidth]{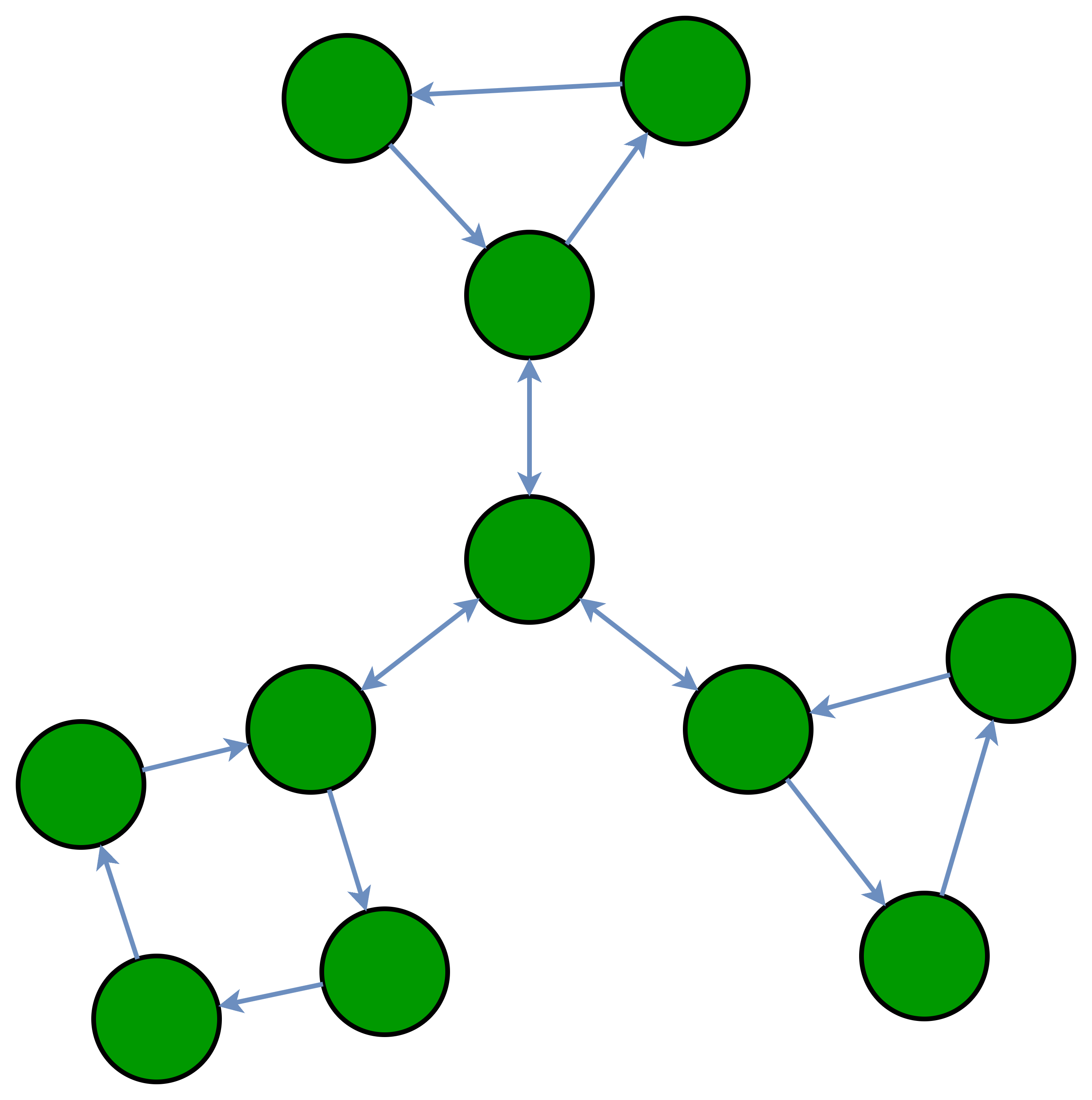}
        \caption{Hybrid}
        \label{fig:hybrid_topo}
    \end{subfigure}
    \caption{Examples of topologies used in distributed systems: Line, Mesh, Star, and Hybrid.}
    \label{fig:topologies}
\end{figure}
\subsection{Decentralised Federated Learning Paradigms}
DFL architectures can be broadly classified into two paradigms: Continual and Aggregate \cite{Mart_nez_Beltr_n_2023} \cite{yuan2024decentralizedfederatedlearningsurvey}. In the Continual paradigm, clients sequentially train a shared model received from another client without intermediate aggregation. This process relies on passing the model from one client to the next for further training. In the Aggregate paradigm, clients train local models independently and periodically aggregate their updates with one or more received models.
These paradigms differ in communication resource requirements and model convergence stability. The Continual paradigm has minimal communication overhead, as only one model is transmitted at a time, and no aggregation computation is required. In contrast, the Aggregate paradigm incurs higher communication costs, as it involves transmitting multiple model parameters and requires an efficient aggregation algorithm \cite{gabrielli2023surveydecentralizedfederatedlearning} 
 \cite{yuan2024decentralizedfederatedlearningsurvey}. Model convergence stability is another key distinction. In the Continual paradigm, as the model sequentially trains across multiple clients, early training information may be forgotten, and the order of client communication can significantly influence convergence. This can lead to instability in the learning process. Conversely, the Aggregate paradigm provides more stable convergence and improved generalisation performance through periodic aggregation, ensuring that contributions from all clients are preserved \cite{gabrielli2023surveydecentralizedfederatedlearning} \cite{yuan2024decentralizedfederatedlearningsurvey}. Choosing between these paradigms involves balancing trade-offs based on deployment scenarios, device performance, and the specific requirements of the task. For instance, the Continual paradigm may be preferable in resource-constrained environments where communication overhead is a critical limitation. On the other hand, the Aggregate paradigm is better suited for scenarios prioritising model stability and performance despite higher communication costs.
 
\textbf{Continual Paradigm }: In the study by Yuan et al., a continuous paradigm was used. They introduced FedPC, a novel peer-to-peer (P2P) FL framework designed for naturalistic driving action recognition. This serverless framework leverages continual learning and a gossip communication protocol to facilitate direct model updates between clients. By eliminating the central server, FedPC reduces communication, computational, and storage overheads, enhancing privacy and learning efficiency. The gossip protocol introduces randomness in communication, improving system robustness and reducing single points of failure. Experimental results demonstrate that FedPC achieves competitive performance compared to traditional client-server FL approaches, with superior knowledge dissemination rates and efficient resource usage, making it suitable for dynamic and heterogeneous environments.

\textbf{Aggregate Paradigm}: Research using the aggregate paradigm has also achieved good results, e.g. Shi et al. present an AirComp-based DSGT-VR (Decentralized Stochastic Gradient Tracking with Variance Reduction) algorithm for decentralized FL over wireless networks \cite{shi2021overtheairdecentralizedfederatedlearning}. This approach eliminates the central parameter server by using device-to-device (D2D) communication and over-the-air computation (AirComp) to manage additive noise and adapt to changes in network topology. The algorithm integrates gradient tracking and variance reduction techniques to enhance convergence performance, ensuring linear convergence despite channel fading and noise. Key features include precoding and decoding strategies to support reliable D2D communication and effective model aggregation. Numerical experiments validate the algorithm's superior performance in terms of convergence rate and accuracy, highlighting its potential in scenarios where central coordination is impractical. In the MATCHA algorithm proposed by Wang et al., a more novel Dynamic parallel broadcast-gossip hybrid topology is used, which optimizes communication in decentralized stochastic gradient descent \cite{9944194}. The algorithm employs a matching-based link scheduling strategy, decomposing the network topology into several matching of non-overlapping links. This ensures nodes communicate only with a subset of neighbours in each iteration, reducing communication overhead while maintaining connectivity. MATCHA optimizes activation probabilities of these matchings, balancing connectivity and efficiency, and pre-computes a sequence of random topologies to enhance training efficiency. Experimental results show that MATCHA achieves significant reductions in training time while maintaining or improving model accuracy, marking a notable advancement in decentralized training efficiency. Each method introduces unique characteristics to optimize decentralized federated learning. FedPC leverages continual learning and a random gossip protocol to enhance privacy and efficiency in a P2P framework. The AirComp-based DSGT-VR algorithm integrates gradient tracking and variance reduction with over-the-air computation to manage communication noise and adapt to network changes in wireless settings. MATCHA employs a sophisticated link scheduling strategy to optimize communication among nodes, focusing on balancing connectivity and efficiency without a central server. 
\subsection{Parallel Stochastic Gradient Descent (SGD)}
Stochastic Gradient Descent (SGD) is a widely used optimisation algorithm in machine learning, particularly for optimising smooth objective functions. Unlike traditional optimisation methods, SGD updates model parameters iteratively using small, randomly sampled subsets of data (mini-batches), making it computationally efficient for large-scale problems. Research by Moulines et al. and Nemirovski et al. established that SGD achieves a convergence rate of $ O(1 / \sqrt{K}) $, for general convex problems, where K refers to the number of iterations for the convergence rate and $ O(1 / K) $ for strongly convex problems \cite{NIPS2011_40008b9a} \cite{RobustStochasticApproximation}. In Collaborative Federated Learning (CFL), Centralized Parallel SGD (C-PSGD) extends the traditional SGD algorithm to leverage parallelism, significantly enhancing training speed in distributed systems. During training, each client independently updates its local model using SGD. The central server then collects and aggregates these local models to update the global model.
In their study on asynchronous decentralised parallel stochastic gradient descent (PSGD), Lian et al. demonstrated that, with appropriate tuning of learning rates and gradient aggregation methods, the errors introduced by parallelism can be effectively controlled. This enables parallel SGD to achieve performance comparable to traditional SGD, with a convergence rate of $ O(1 / \sqrt{K*n}) $,where n represents the number of participating clients.\cite{lian2018asynchronousdecentralizedparallelstochastic}.

Decentralised Parallel Stochastic Gradient Descent (D-PSGD) extends Centralized Parallel SGD by eliminating the need for a central server. Instead, devices perform peer-to-peer communication to exchange and aggregate updated parameters. Lan et al. provided a foundational theoretical analysis demonstrating that D-PSGD can outperform C-PSGD under certain conditions, a claim validated through experimental results \cite{lian2017can}.

D-PSGD is generally implemented in two forms: synchronous and asynchronous. In synchronous D-PSGD, devices must wait for all peers to complete local training before simultaneously communicating and aggregating updates. In contrast, asynchronous D-PSGD allows devices to independently update their local model parameters without waiting for others. Studies by Ram et al., Srivastava and Nedic, and Sundhar Ram et al. have examined the convergence properties of these approaches \cite{ram2008distributedstochasticsubgradientprojection} \cite{8483382} \cite{5719290}.
Asynchronous D-PSGD typically enables faster training by reducing idle time but introduces added complexity and potential challenges in guaranteeing convergence. For this reason, we opted to use synchronous D-PSGD in our experimental design, prioritising stable convergence and facilitating reliable performance evaluation.
\section{Methodology} 
This research proposes a robust DFL framework to improve model convergence and address communication and computational challenges on resource-constrained, dynamically connected edge devices\cite{10167739}, with convergence achieved using Decentralised Parallel Stochastic Gradient Descent with random gossip protocols(D-PSGD). 
Given the limited communication bandwidth of edge devices, which makes one-to-many broadcast communication infeasible, this research employs a random 1-to-1 gossip protocol as the primary communication method. This protocol ensures communication occurs between two devices per round, effectively addressing bandwidth constraints while enhancing scalability and robustness. Although D-PSGD with random gossip protocols may converge more slowly than centralised SGD due to decentralised communication, it demonstrates significant advantages in large-scale, heterogeneous networks.

The dynamic nature of edge environments adds complexity to deployment. Devices mounted on mobile platforms (e.g., vehicles or animals) frequently enter and leave communication ranges, resulting in constantly shifting network topologies. The random gossip protocol, which randomly selects a communication partner in each round, introduces substantial flexibility for such dynamic settings but also creates challenges.
The dynamic nature of communication ranges, represented as \( r \), leads to continually shifting network topologies, where \( G(V, E) \) represents the graph formed by devices \( V \) and communication links \( E \). The random gossip protocol, denoted as \( P_t \), selects a communication partner \( v_j \in V \) for a device \( v_i \in V \) at each round \( t \) with probability proportional to adjacency \( A_{ij} \), introducing flexibility but also computational complexity in these dynamic settings.

The convergence rate of D-PSGD with random gossip protocols is dependent on the \textit{spectral gap} of the Laplacian matrix \( L \) of \( G \), given as \( \rho = 1 - \lambda_2(L) \), where:
\begin{itemize}
    \item \( \lambda_2(L) \) is the second smallest eigenvalue of \( L \),
    \item A larger spectral gap \( \rho \) (or equivalently a smaller \( \lambda_2 \)) indicates better connectivity and facilitates faster convergence of the protocol.
\end{itemize}

Thus, the convergence speed \( T_{\text{convergence}} \) is inversely related to the spectral gap:
\[
T_{\text{convergence}} \propto \frac{1}{\rho} = \frac{1}{1 - \lambda_2}.
\]
A well-connected graph \( G \) with a larger \( \rho \) leads to efficient information dissemination and faster convergence of the algorithm \cite{pmlr-v97-koloskova19a}.

Conversely, sparse or poorly connected networks slow information propagation, impeding convergence, while fully connected or dense topologies achieve superior performance. Alexandros et al. demonstrated that random 1-to-1 gossip underperforms on grids and random geometric graphs, where information diffusion resembles a random walk \cite{5545370}. For instance, in a two-dimensional grid with side length \( d \), it requires \( d^2 \) steps to traverse the network entirely. Introducing directionality in information propagation significantly accelerates diffusion within random gossip protocols.

\begin{figure}[h!]
  \centering
    \begin{subfigure}[t]{0.40\linewidth}
        \centering
        \includegraphics[width=\linewidth]{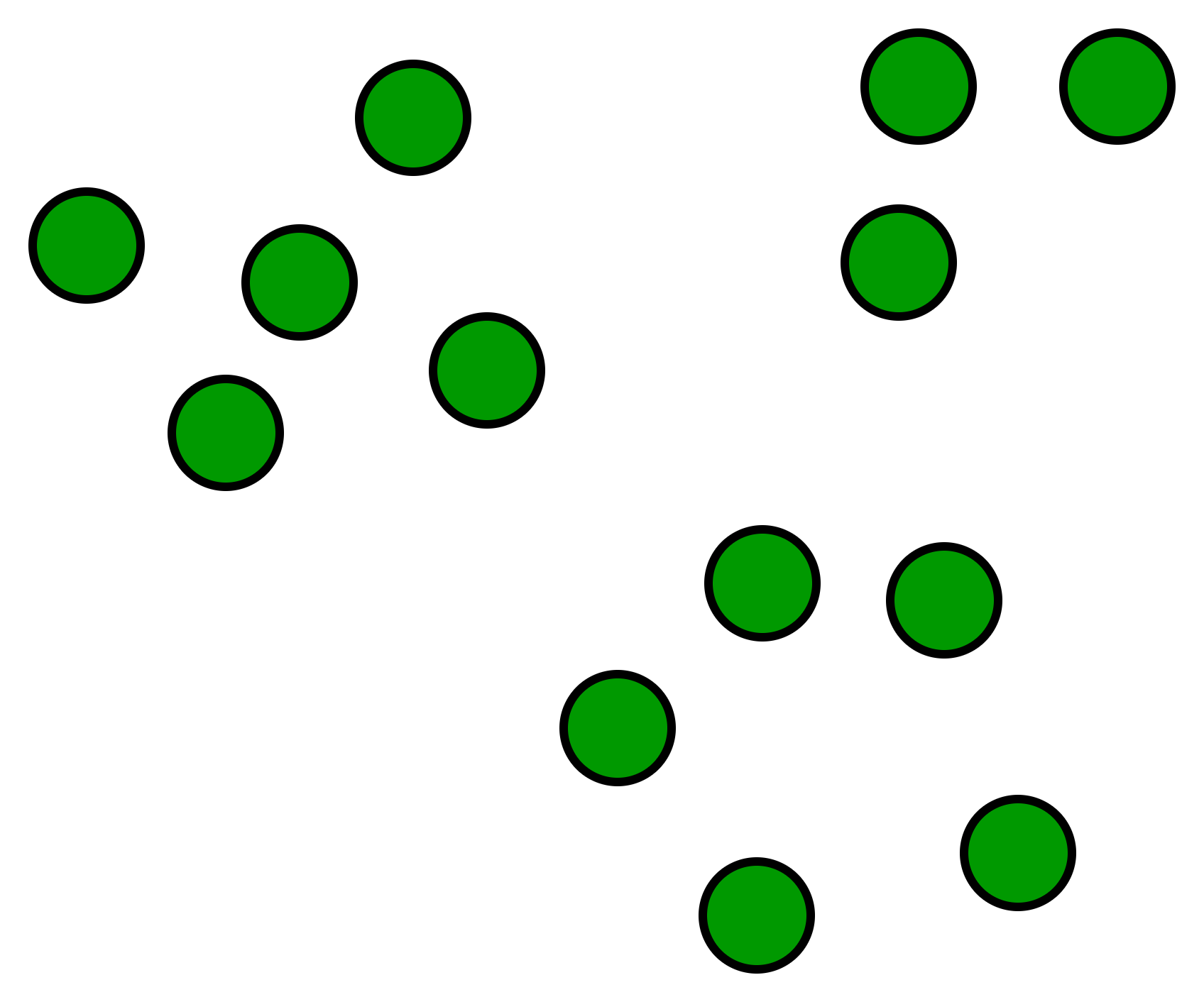}
        \caption{Before DK-means.}
        \label{fig:f1}
    \end{subfigure}
    \hfill
    \begin{subfigure}[t]{0.40\linewidth}
        \centering
        \includegraphics[width=\linewidth]{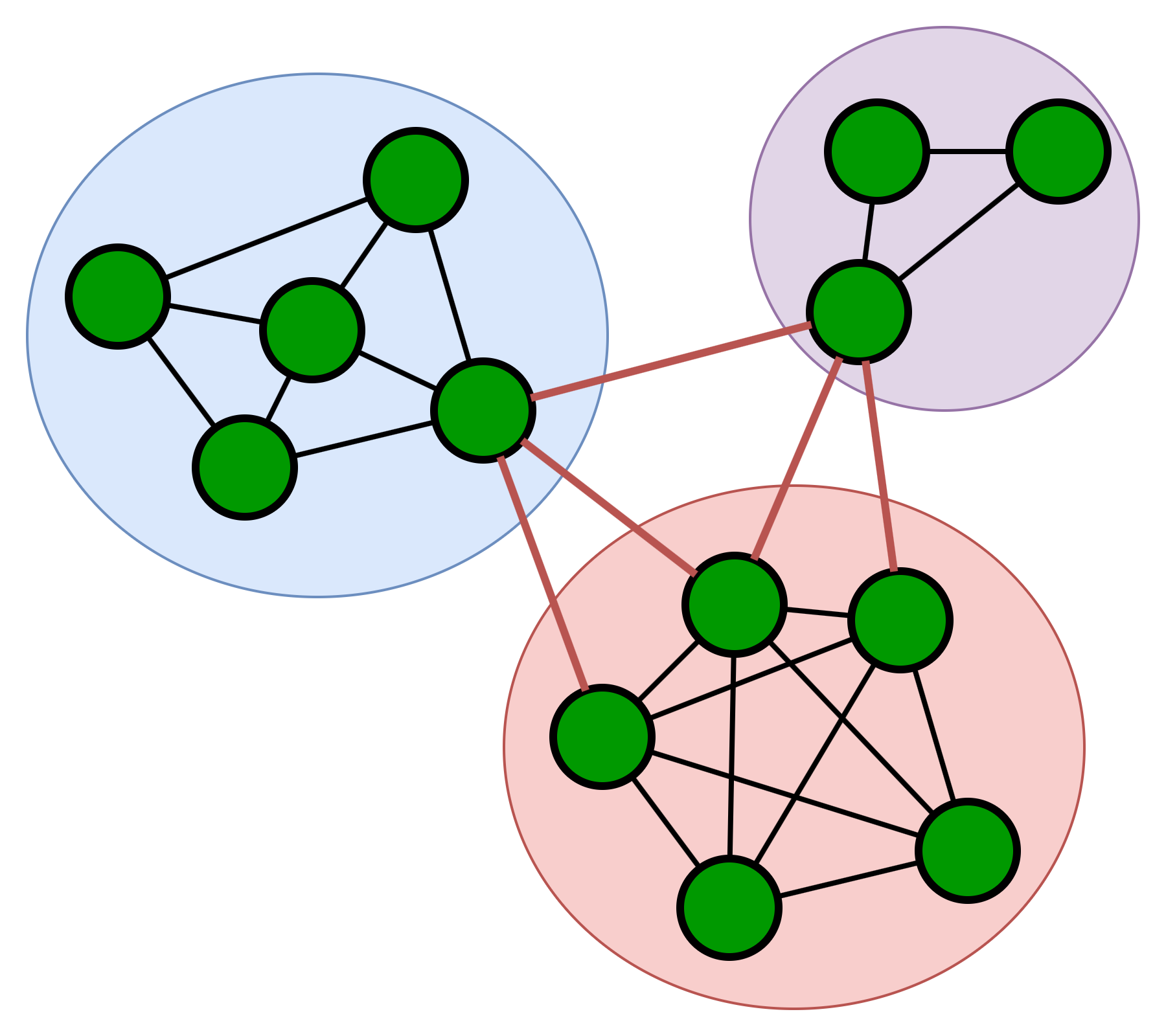}
        \caption{After DK-means.}
        \label{fig:f2}
    \end{subfigure}
    \caption{(a) Devices before DK-means, and (b) the topology after DK-means. Black edges represent intra-cluster topology, and red edges represent inter-cluster topology. Due to limited communication range, inter-cluster topology is sparser.}
    \label{fig:dkmeans}
\end{figure}

To address these limitations, this research employs the Decentralized K-means (DK-means) algorithm to group devices into clusters based on geographical proximity. The proposed framework introduces a hierarchical two-layer topology: intra-cluster and inter-cluster layers. Within each cluster, devices communicate using the gossip protocol, facilitating information exchange and local model aggregation. Since devices within a cluster are geographically close, their network is denser and approximates a fully connected topology, enabling faster convergence of the gossip protocol (see Figure~\ref{fig:dkmeans}).

\begin{figure}[h!]
  \centering
    \begin{subfigure}[t]{0.40\linewidth}
        \centering
        \includegraphics[width=\linewidth]{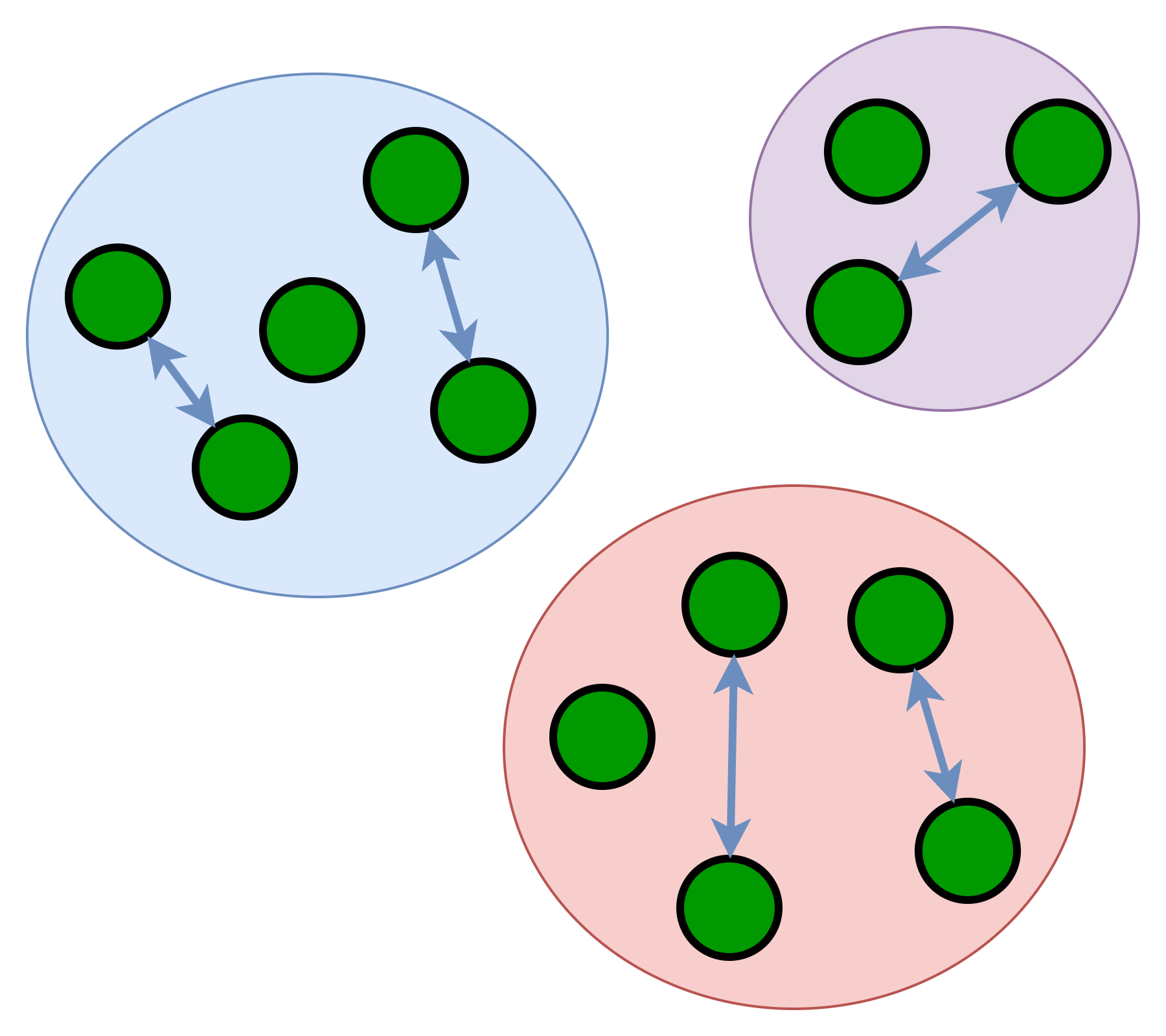}
        \caption{Intra-cluster gossip.}
        \label{fig:f3}
    \end{subfigure}
    \hfill
    \begin{subfigure}[t]{0.40\linewidth}
        \centering
        \includegraphics[width=\linewidth]{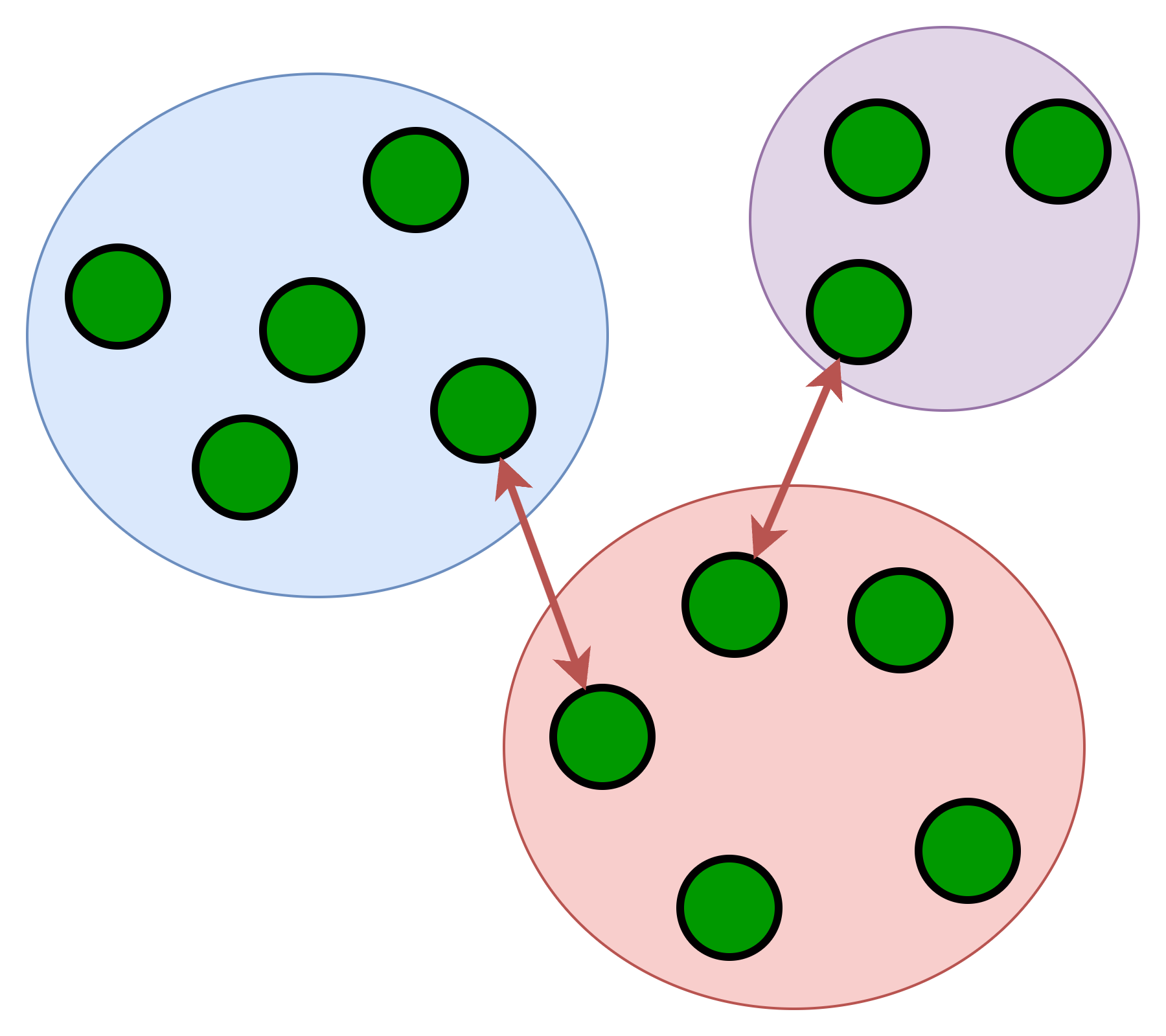}
        \caption{Inter-cluster gossip.}
        \label{fig:f4}
    \end{subfigure}
    \caption{(a) Devices exchange weights via 1-to-1 gossip protocol within clusters, and (b) inter-cluster communication. Devices prioritise neighbours not previously contacted, increasing efficiency in weight dissemination.}
    \label{fig:gossip}
\end{figure}

The inter-cluster layer, in contrast, is sparser. However, because model aggregation has already occurred within clusters, devices only need to communicate with those in other clusters that fall within their communication range. This hierarchical structure leverages dense intra-cluster communication for rapid convergence and reduces inter-cluster communication overhead, ensuring efficient information diffusion across the entire network. This design effectively balances the trade-offs between scalability, resource constraints, and convergence performance (see Figure~\ref{fig:gossip}).

\section{Algorithms}

\begin{algorithm}
\caption{D-PSGD with 1-to-1 Gossip Protocol}
\begin{algorithmic}[1]
\State \textbf{Input:} Initial node states \( \mathbf{x}_i \in \mathbb{R}^d, i \in \{1, 2, \dots, n\} \),
\State Number of intra-cluster gossip iterations \( G_{\text{intra}} \),
\State Number of inter-cluster gossip iterations \( G_{\text{inter}} \),
\State Total number of iterations \( K \).

\vspace{0.5em}
\State \textbf{Step 1: Cluster Formation}
\State Use DK-means to form \( m \) clusters.

\vspace{0.5em}
\State \textbf{Step 2: Local Updates}
\State Each node \( i \) performs local updates in parallel:
\[
\mathbf{x}_i^{(k+1)} = \mathbf{x}_i^{(k)} - \eta \nabla f_i(\mathbf{x}_i^{(k)}),
\]

\vspace{0.5em}
\State \textbf{Step 3: Intra-cluster Gossip Communication}
\State \textcolor{blue}{\vrule height 9pt width 1pt depth 1pt \kern 4pt} \textbf{For} $g = 0, 1, 2, \dots, G_{\text{intra}} - 1$
\State \textcolor{blue}{\vrule height 9pt width 1pt depth 1pt \kern 8pt} Randomly select a neighbouring node \( j \in \mathcal{C}_c \)
\State \textcolor{blue}{\vrule height 9pt width 1pt depth 1pt \kern 8pt} Exchange model weights between nodes \( i \) and \( j \):
\[
\mathbf{x}_i^{(g+1)} = \alpha \mathbf{x}_i^{(g)} + (1-\alpha) \mathbf{x}_j^{(g)}.
\]
\State \textcolor{blue}{\vrule height 9pt width 1pt depth 1pt \kern 8pt} Perform cumulative FedAvg for model aggregation.
\State \textcolor{blue}{\vrule height 9pt width 1pt depth 1pt \kern 4pt} \textbf{End For}

\vspace{0.5em}
\State \textbf{Step 4: Inter-cluster Gossip Communication}
\State \textcolor{blue}{\vrule height 9pt width 1pt depth 1pt \kern 4pt} \textbf{For} $g = 0, 1, 2, \dots, G_{\text{inter}} - 1$
\State \textcolor{blue}{\vrule height 9pt width 1pt depth 1pt \kern 8pt} Randomly select a neighbouring node \( j \notin \mathcal{C}_c \)
\State \textcolor{blue}{\vrule height 9pt width 1pt depth 1pt \kern 8pt} Exchange model weights between nodes \( i \) and \( j \):
\[
\mathbf{x}_i^{(g+1)} = \alpha \mathbf{x}_i^{(g)} + (1-\alpha) \mathbf{x}_j^{(g)}.
\]
\State \textcolor{blue}{\vrule height 9pt width 1pt depth 1pt \kern 8pt} Perform FedAvg for inter-cluster aggregation.
\State \textcolor{blue}{\vrule height 9pt width 1pt depth 1pt \kern 4pt} \textbf{End For}

\end{algorithmic}
\label{alg:dpsgd}
\end{algorithm}

The algorithm builds on the principles previously described, utilising the DK-means algorithm to construct a bilayer network, which comprises:
\begin{itemize}
    \item \textbf{Intra-cluster layer:} A densely connected network within each cluster.
    \item \textbf{Inter-cluster layer:} A sparse network between clusters.
\end{itemize}

At the start of each round, the DK-means algorithm dynamically forms clusters based on device locations. Communication within clusters is conducted over $G_{\text{intra}}$ iterations using a 1-to-1 gossip protocol, see algorithm\ref{alg:dpsgd}. Cumulative FedAvg is employed for model aggregation to minimise storage requirements on edge devices. Following intra-cluster communication, the process is repeated for $G_{\text{inter}}$ iterations on the inter-cluster layer, completing one training round.

The standard K-means algorithm requires all data to be centralised, which is impractical without a central server. The Distributed K-means (DK-means) in algorithm \ref{alg:DKmean} operates without data centralisation by enabling each node to perform computations locally and communicate with others to achieve global clustering.

\begin{algorithm}
\caption{Distributed K-means Algorithm}
\label{alg:kmeans}
\begin{algorithmic}[1]
\State \textbf{Inputs:} Number of clusters \( K \), set of devices \( D = \{d_1, d_2, \dots, d_n\} \), communication range \( r \).
\State Randomly initialise \( K \) cluster heads \( C = \{c_1, c_2, \dots, c_K\} \) from \( D \).

\vspace{0.5em}
\Repeat
    \State \textcolor{blue}{\vrule height 9pt width 1pt depth 1pt \kern 4pt} \textbf{For each device} \( d_i \notin C \): 
    \State \textcolor{blue}{\vrule height 9pt width 1pt depth 1pt \kern 8pt} Search for the nearest cluster head \( c_j \) within  \( r \): 
    \[
    c_j = \arg\min_{c \in C} \|d_i - c\|
    \]
    \State \textcolor{blue}{\vrule height 9pt width 1pt depth 1pt \kern 8pt} Assign \( d_i \) to the cluster of \( c_j \): 
    \[
    \mathcal{C}_j \gets \mathcal{C}_j \cup \{d_i\}.
    \]
    \State \textcolor{blue}{\vrule height 9pt width 1pt depth 1pt \kern 8pt} \textbf{If} no cluster head is found:
    \State \textcolor{blue}{\vrule height 9pt width 1pt depth 1pt \kern 12pt} \textbf{If} \( |N(d_i)| \geq 1 \), \( N(d_i) \) =set of devices within \( r \):
    \State \textcolor{blue}{\vrule height 9pt width 1pt depth 1pt \kern 16pt} Join the cluster of the nearest device \( d_k \):
    \[
    \mathcal{C}_k \gets \mathcal{C}_k \cup \{d_i\}.
    \]
    \State \textcolor{blue}{\vrule height 9pt width 1pt depth 1pt \kern 12pt} \textbf{Else}:
    \State \textcolor{blue}{\vrule height 9pt width 1pt depth 1pt \kern 16pt} Device \( d_i \) becomes a new cluster head:
    \[
    C \gets C \cup \{d_i\}.
    \]

    \State \textcolor{blue}{\vrule height 9pt width 1pt depth 1pt \kern 4pt} \textbf{For each cluster head} \( c_j \in C \):
    \State \textcolor{blue}{\vrule height 9pt width 1pt depth 1pt \kern 8pt} Store IDs and locations of devices in \( \mathcal{C}_j \). 
    \State \textcolor{blue}{\vrule height 9pt width 1pt depth 1pt \kern 8pt} \textbf{If} \( |\mathcal{C}_j| > 0 \):
    \State \textcolor{blue}{\vrule height 9pt width 1pt depth 1pt \kern 12pt} Update \( c_j \):
    \[
    c_j = \arg\min_{d \in \mathcal{C}_j} \|d - \bar{d}_j\|, \quad \bar{d}_j = \frac{1}{|\mathcal{C}_j|} \sum_{d \in \mathcal{C}_j} d.
    \]
    \State \textcolor{blue}{\vrule height 9pt width 1pt depth 1pt \kern 8pt} \textbf{Else}:
    \State \textcolor{blue}{\vrule height 9pt width 1pt depth 1pt \kern 12pt} Update \( c_j \) to another random device:
    \[
    c_j = \text{random}(D).
    \]
\Until{Cluster head positions stabilise or maximum iterations \( T_{\max} \) is reached.}

\State \textbf{Output:} Final clusters \( \mathcal{C}_1, \mathcal{C}_2, \dots, \mathcal{C}_K \) \& heads \( C \).
\end{algorithmic}
\label{alg:DKmean}
\end{algorithm}

Algorithm\ref{alg:Cumulative}, is a memory-efficient variant of the standard FedAvg algorithm. Its design addresses the limited memory capacity of edge devices. During the communication phase, devices avoid storing multiple models locally for aggregation. Instead, each device calculates a cumulative weighted sum of received models and keeps track of the total number of samples used to train them. This approach requires only the memory space for a single model. After communication ends, the weighted sum is divided by the total number of samples to produce the aggregated model.

\begin{algorithm}
\caption{Cumulative FedAvg Algorithm}
\label{alg:fedavg}
\begin{algorithmic}[1]
\State \textbf{Inputs:}
\State \hspace{\algorithmicindent} \( W \): New weight vector from the gossip protocol 
\State \hspace{\algorithmicindent} \( W_{\text{sum}} \): Weighted sum of all weights already received 
\State \hspace{\algorithmicindent} \( w \): Individual weight in \( W \) 
\State \hspace{\algorithmicindent} \( N \): Number of samples used to train \( W \) 
\State \hspace{\algorithmicindent} \( N_{\text{sum}} \): Total number of samples from all weights

\vspace{0.5em}
\State \textbf{During Gossip Communication:}
\State \( W_{\text{result}} = \emptyset \)
\State \textcolor{blue}{\vrule height 9pt width 1pt depth 1pt \kern 4pt} \textbf{For each pair} \( w_1 \in W_{\text{sum}}, w_2 \in W \):
\State \textcolor{blue}{\vrule height 9pt width 1pt depth 1pt \kern 8pt} Update the cumulative weighted sum of weights:
\[
w_{\text{combined}} = w_1 + w_2 \cdot N
\]
\State \textcolor{blue}{\vrule height 9pt width 1pt depth 1pt \kern 8pt} Append \( w_{\text{combined}} \) to \( W_{\text{result}} \)
\State \textcolor{blue}{\vrule height 9pt width 1pt depth 1pt \kern 4pt} \textbf{End For}
\State Update the total sample count:
\[
N_{\text{sum}} = N_{\text{sum}} + N
\]
\State Update the cumulative weight sum:
\[
W_{\text{sum}} = W_{\text{result}}
\]
\vspace{0.5em}
\State \textbf{After Gossip Communication:}
\State \( W_{\text{aggregated}} = \emptyset \)
\State \textcolor{blue}{\vrule height 9pt width 1pt depth 1pt \kern 4pt} \textbf{For each} \( w \in W_{\text{sum}} \):
\State \textcolor{blue}{\vrule height 9pt width 1pt depth 1pt \kern 8pt} Compute the final aggregated weight:
\[
w_{\text{temp}} = \frac{w}{N_{\text{sum}}}
\]
\State \textcolor{blue}{\vrule height 9pt width 1pt depth 1pt \kern 8pt} Append \( w_{\text{temp}} \) to \( W_{\text{aggregated}} \)
\State \textcolor{blue}{\vrule height 9pt width 1pt depth 1pt \kern 4pt} \textbf{End For}

\State \Return \( W_{\text{aggregated}} \)
\end{algorithmic}
\label{alg:Cumulative}
\end{algorithm}

\section{Convergence Framework}

Consider an undirected network \( G:(V,W) \) with \( n \) worker nodes. \( V \) is a set containing all worker nodes: \( V:\{1,2,\dots,n\} \). \( W \in \mathbb{R}^{n \times n} \) is a symmetric weight matrix, indicates the connectivity between devices, where \( W_{ij} = W_{ji}\), and \( W_{ij} \in [0,1] \), with 0 indicating that there is no connection between device \( i \) and device \( j \), and 1 indicating the opposite. \( D_i: \{(x_i^1, y_i^1), (x_i^2, y_i^2), \dots, (x_i^d, y_i^d)\} \) represents the local dataset of device \( i \) with size $d$.
\textbf{Assumptions:} Before we start analysing the convergence rate of D-PSGD, the commonly used assumptions are as follows:

\begin{itemize}
    \item \textbf{Lipschitz continuity of the gradient}: All functions \( f_i(\cdot):\mathbb{R}^{d} \rightarrow \mathbb{R} \textit{ for } i \in [n] \) have gradients that adhere to an L-Lipschitz condition.
    \item \textbf{Convexity}: All functions \( f_i(\cdot) \) are µ-strongly convex.
    \item \textbf{Spectral gap}: Given a symmetric stochastic matrix \( W \), we define the spectral gap parameter \( \rho \) as \( \left(\max \left\{\left|\lambda_2(W)\right|,\left|\lambda_n(W)\right|\right\}\right)^{2} \). Here, \( \lambda_2(W) \) and \( \lambda_n(W) \) are the second largest and smallest eigenvalues of \( W \), respectively. We assume \( \rho < 1 \).
    \item \textbf{Bounded variance of the stochastic gradient}: We assume that the variance of the stochastic gradient \( \mathbb{E}_{i \sim \mathcal{U}([n])} \mathbb{E}_{\xi \sim \mathcal{D}_i} \left\| \nabla F_i(x; \xi) - \nabla f(x) \right\|^2 \) is bounded for any node \( x \), where \( i \) is uniformly sampled from \( V:\{1, \ldots, n\} \) and \( \xi \) is sampled from the distribution \( \mathcal{D}_i \). This implies the existence of constants \( \sigma \) and \( \varsigma \) such that:
\[
\mathbb{E}_{\xi \sim \mathcal{D}_i} \left\| \nabla F_i(x; \xi) - \nabla f_i(x) \right\|^2 \leq \sigma^2, \quad \forall i, \forall x,
\]
\[
\mathbb{E}_{i \sim \mathcal{U}([n])} \left\| \nabla f_i(x) - \nabla f(x) \right\|^2 \leq \varsigma^2, \quad \forall x.
\]
\end{itemize}
In a typical distributed optimization problem, the \textbf{objective function} can be written as:

\[
\min_{x \in \mathbb{R}^N} f(x) = \frac{1}{n} \sum_{i=1}^{n} \underbrace{\mathbb{E}_{\xi \sim \mathcal{D}_i} F_i(x; \xi)}_{=: f_i(x)}
\]

where \( F_i(x; \xi) \) is the SGD loss function at node \( i \) which depends on the model parameter \( x \).

\subsection{Convergence Rate of D-PSGD}
Based on existing research, we can find that under the above assumptions, the convergence rate of D-PSGD can be bounded as:

\textbf{\textit{Theorem 1:}}
\[
\frac{\sum_{t=0}^{T-1} \mathbb{E}\left\|\nabla f\left(\frac{X_t \mathbf{1}_n}{n}\right)\right\|^{2}}{T} \leq 
\]
\[
\frac{8(f(0) - f^*)L}{T} + \frac{(8(f(0) - f^*) + 4L)\sigma}{\sqrt{Tn}},
\]
where \( T \) is the total number of iterations, \( n \) is the number of nodes, \( f(0) \) and \( f^* \) are the initial and optimal objective function values, \( L \) is the Lipschitz constant, and \( \sigma \) is a constant bound on the variance of the stochastic gradients as in the assumptions \cite{lian2017can}. Therefore, we can derive the convergence rate as \( \mathcal{O}(1/T + 1/\sqrt{nT}) \). It is worth mentioning that the generally recognized convergence rate of C-PSGD is \( \mathcal{O}(1/T + 1/\sqrt{T}) \) when the objective function f(x) is strongly convex, which shows that the convergence rate of D-PSGD can reach the linear speedup of C-PSGD. 

\subsection{Convergence Rate of D-PSGD with Gossip Protocol}
Based on the existing research we know the conclusion that the averaging time of the distributed synchronous gossip algorithm \( T_{\mathrm{ave}}(\epsilon) \) is bounded by:

\textbf{\textit{Theorem 2:}}
\[
\frac{0.5 \log \epsilon^{-1}}{\log \lambda^{-1}} \leq T_{\mathrm{ave}}(\epsilon) \leq \frac{3 \log \epsilon^{-1}}{\log \lambda^{-1}},
\]
where \( \epsilon \) is the accuracy and \( \lambda \) is the second-largest eigenvalue of \( W \) \cite{1638541}.

Since the averaging time \( T_{\mathrm{ave}}(\epsilon) \) directly affects the convergence rate of the D-PSGD algorithm, specifically, the averaging time reflects how quickly the nodes reach agreement, which in turn influences the overall algorithm's convergence. From the above theorem, we can know that, for any positive \( \epsilon \), after \( T_{\mathrm{ave}}(\epsilon) \) iterations, the nodes achieve an \( \epsilon \)-consensus. Therefore, the convergence rate of D-PSGD will be influenced by this bound.

The basic form of the convergence rate for D-PSGD with gossip that can achieve an \( \epsilon \)-consensus can be approximated as:
\[
\mathbb{E}\left[ f\left(\bar{\mathbf{w}}^T \right) - f(\mathbf{w}^*) \right] \leq O\left( \frac{1}{T} + \frac{1}{\sqrt{nT}} + E_{\text{cons}}\right)
\]
where \(T\) is the total number of iterations and \(\bar{\mathbf{w}}^T\) represents the averaged parameter vector after \(T\) iterations. \(E_{\text{cons}} \) represents the consensus error term, which indicates the extra convergence time caused by gossip protocol.

The \textbf{\textit{Theorem 2}} suggests that the time $T_{\mathrm{ave}}$ required to achieve a specific error $\epsilon$ is logarithmically dependent on the inverse of $\epsilon$:

\begin{align*}
    T_{\mathrm{ave}}(\epsilon) &\approx c \cdot \frac{\log \epsilon^{-1}}{\log \lambda_{\max}^{-1}} \quad \text{(from \textbf{Theorem 2})}, \\
    \epsilon &\approx \exp \left( - c \cdot T_{\mathrm{ave}} \log \lambda_{\max}^{-1} \right) \quad (\text{by exponentiation}).
\end{align*}

For an appropriately chosen constant $c$, we can derive:
\[
E_{\text{cons}} = O\left( \exp \left( - c \cdot T_{\mathrm{ave}} \log \lambda_{\max}^{-1} \right) \right)
\]
\[
E_{\text{cons}} = O\left( \exp \left( c \cdot T_{\mathrm{ave}} \log \lambda_{\max} \right) \right).
\]

Combining the averaging time bound and convergence rate bound, we derive the following:
\textbf{\textit{Theorem 3:}}
\[
\begin{aligned}
\mathbb{E}\left[ f\left(\bar{\mathbf{w}}^T \right) - f\left(\mathbf{w}^*\right) \right] &\leq \\
O \left( \frac{1}{T} + \frac{1}{\sqrt{nT}} + \exp \left( c \cdot T_{\mathrm{ave}} \log \lambda_{\max} \right) \right).
\end{aligned}
\]

Therefore, the convergence rate for D-PSGD with gossip simplifies to:
\[
O \left( \frac{1}{T} + \frac{1}{\sqrt{nT}} + \exp \left( c \cdot T_{\mathrm{ave}} \log \lambda_{\max} \right) \right).
\]

This implies that a smaller $\lambda_{\max}$ (or larger spectral gap $1 - \lambda_{\max}$) results in faster convergence to consensus, consistent with the findings of Koloskova et al.'s study \cite{pmlr-v97-koloskova19a}.

\subsection{The Overall Convergence Rate of D-PSGD with 1-to-1 Gossip Protocol on a Bilayer Network}

After performing the intra-cluster D-PSGD, devices must communicate with devices from other clusters. This creates a sparser network, where the spectral properties of the inter-cluster communication matrix differ. Specifically, the second-largest eigenvalue $\lambda'_{\max}$ is typically larger, leading to slower convergence for achieving consensus across clusters.

The overall convergence rate is influenced by both the intra-cluster and inter-cluster phases. These phases are independent, so the total complexity is the sum of the complexities of the two phases. Let the spectral properties and averaging times for the networks be defined as follows:

\begin{itemize}
    \item $\lambda_{\max}^{\text{intra}}$: The second-largest eigenvalue of the intra-cluster communication matrix.
    \item $\lambda_{\max}^{\text{inter}}$: The second-largest eigenvalue of the inter-cluster communication matrix.
    \item $T_{\mathrm{ave}}^{\text{intra}}$: The averaging time for intra-cluster communication.
    \item $T_{\mathrm{ave}}^{\text{inter}}$: The averaging time for inter-cluster communication.
\end{itemize}

The overall convergence rate can then be expressed as:
\begin{align*}
O \biggl( \frac{1}{T} + \frac{1}{\sqrt{nT}} + \exp \bigl( T_{\mathrm{ave}}^{\text{intra}} \log \lambda_{\max}^{\text{intra}} \bigr) \\
+ \exp \bigl( T_{\mathrm{ave}}^{\text{inter}} \log \lambda_{\max}^{\text{inter}} \bigr) \biggr).
\end{align*}

\section{Experiments}
We will use normal local training and Centralized Federated Learning (CFL) as baselines to evaluate the method we propose. This is because, in terms of model performance,  normal local training is widely considered to represent the optimal result, which could serve as an upper bound because it's not limited by the consensus constraints or the communication overhead inherent in decentralized methods. Additionally, the performance of DFL is often compared to CFL, with CFL being the theoretical upper bound of what can be achieved when decentralization constraints are removed. If DFL achieves performance close to that of CFL, it indicates that the DFL method is effectively coordinating the learning process across the network. We will also compare the convergence rate of our proposed DFL method with that of CFL to verify whether DFL can achieve the theoretically expected linear 
Since our goal is to implement a DFL framework that can be deployed on edge devices, we first constructed a simulation environment that incorporates the characteristics of edge devices (figure~\ref{fig:cifar-10}). We used the `pyglet` library to build the foundation of the simulation environment and to visualise the states of the device. The simulated field is a square area of size \(100 \times 100\), within which devices can move freely. Each device has the following attributes:

\begin{itemize}
    \item \textbf{Position coordinates (x, y)}: We assume that each device knows its location in real time, which corresponds to the GPS functionality in the real world.
    \item \textbf{ID}: Each device has a unique ID.
    \item \textbf{Communication range}: Each device can communicate only with other devices within its communication range \textit{r}, simulating the limited communication range of edge devices in reality.
    \item \textbf{Movement speed}: Each device has a movement speed \textit{s} to simulate devices deployed in vehicles or animals. The direction of movement is random and variable and the movement is continuous and slow. This is to ensure that the devices have enough time to complete at least one full communication.
\end{itemize}

\begin{figure}
\centering
\includegraphics[width = 0.7\hsize]{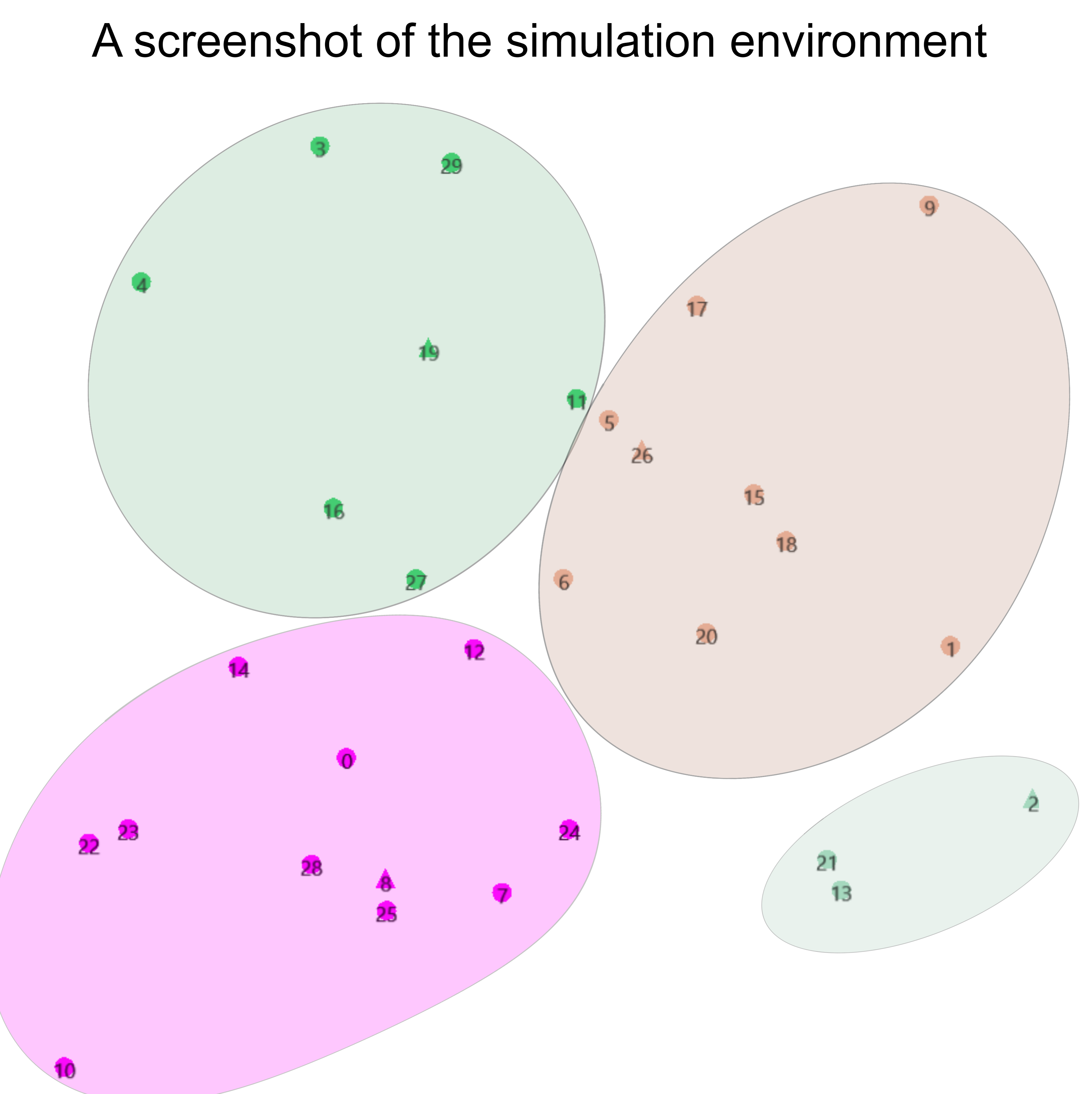}
\caption{Simulation environment with 30 devices. Circles represent normal devices and triangles represent the header of the clusters and the numbers are the unique IDs of the devices. After DK-means the devices are divided into 4 clusters.}
\label{fig:cifar-10}
\end{figure}

The above attributes are related to the simulation environment. Other attributes related to the experiment, such as local models, parameters, and datasets, will be introduced in the next section.
To simulate information transmission between multiple devices on a single computer, rather than simply reading memory from other devices locally, we used the MQTT (Message Queuing Telemetry Transport) library to simulate communication between devices. This is an ISO standard message protocol based on the publish/subscribe paradigm over TCP/IP, specifically designed for remote devices with low hardware performance and under poor network conditions.

\subsection{Dataset and Model}

Our experiment focuses on image recognition datasets, specifically using CIFAR-10. CIFAR-10 consists of 60,000 colour images, each 32x32 pixels, categorised into 10 classes. These classes are evenly split between vehicles (e.g., aeroplanes, ships) and animals (e.g., birds, dogs). Each class contains 6,000 images, ensuring a balanced dataset. We used the official train/test split, with 50,000 images for training and 10,000 for testing \cite{cifar-10}. Examples of the dataset are shown in Figure~\ref{fig:cifar-10}.
\begin{figure}
\centering
    \includegraphics[width=0.7\linewidth]{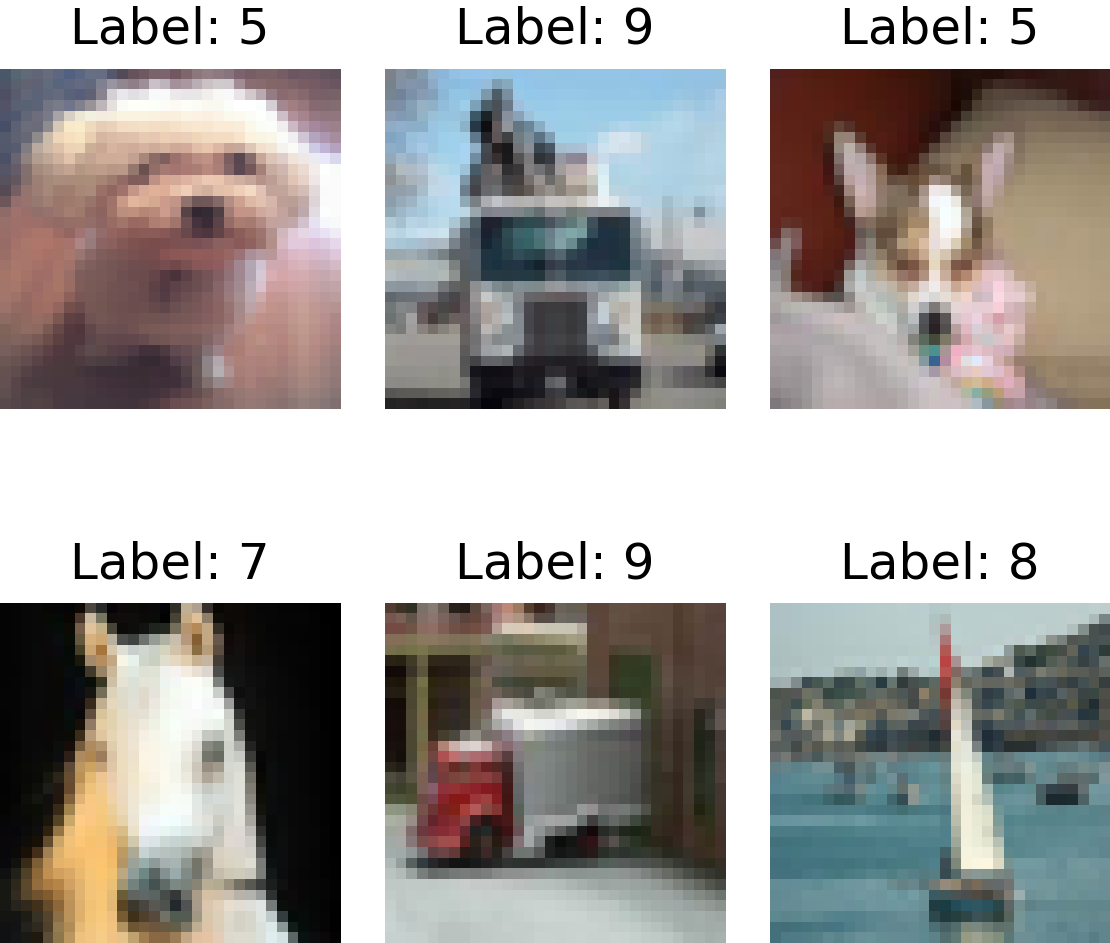}
    \caption{Examples of images in CIFAR-10 dataset.}
    \label{fig:cifar-10}
\end{figure}
In Distributed Federated Learning (DFL), each device is assigned a unique local dataset. For this, we employed two split methods: 1)\textbf{IID Dataset}: In IID datasets, all data points are generated independently from the same probability distribution. This simplifies theoretical analysis and is common in centralised machine learning. Each device receives a unique, balanced subset with an equal number of samples for all classes, 2)\textbf{Non-IID Dataset}: Real-world datasets are typically Non-IID, where devices collect data specific to their environment. For example, a device may only have data from 1-2 classes, or the class distribution may be highly imbalanced. 
The degree of Non-IIDness is controlled by the Dirichlet parameter $\alpha$. A larger $\alpha$ results in a more uniform distribution, while a smaller $\alpha$ creates a skewed distribution. For instance, with $\alpha=0.5$, the dataset is highly imbalanced.

\begin{figure}
  \centering
{\includegraphics[width=0.45\textwidth]{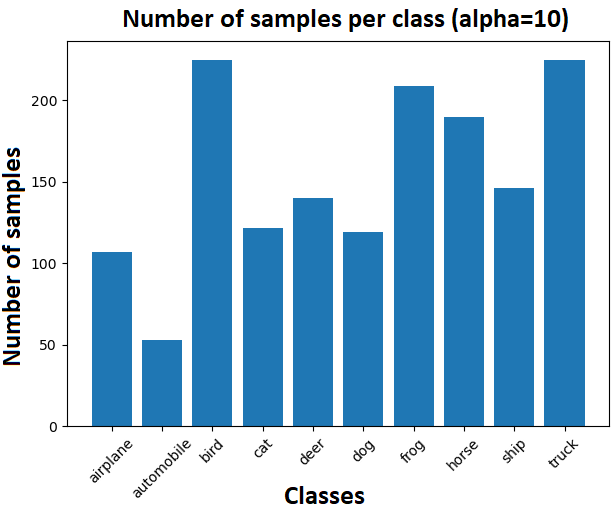}}
{\includegraphics[width=0.45\textwidth]{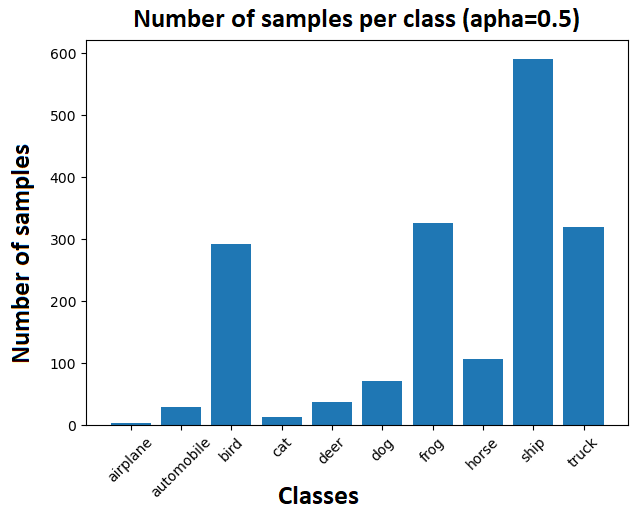}}
  \caption{Label distribution for different $\alpha$ values. Smaller $\alpha$ results in a more skewed class distribution.}
\end{figure}

Due to limited computational resources on edge devices, full-model training is often infeasible. For our experiment, we selected the MCUNet-in3 model, developed by MIT for resource-constrained edge devices. The model has 0.74M parameters and achieves 62.2\% top-1 accuracy and 84.5\% top-5 accuracy on ImageNet.
Given that most microcontrollers (MCUs) support transfer learning but not full-model fine-tuning, we adopted a transfer learning approach. Initially, we performed full-model fine-tuning on the complete training set. Then, we replaced the model's classifier layer with three fully connected layers. In subsequent training, all parameters except the classifier layer were frozen. This significantly reduced computational requirements and increased training speed.
\subsection{Experimental Setup}
The experiment evaluates the performance of two baselines and the proposed bilayer GD-PSGD (bilayer Gossip D-PSGD) method. The two baselines are normal local training and training using the CFL framework. The following settings are consistent across all training setups:
\begin{itemize}
    \item GPU: RTX 3070 Ti.
    \item Framework: PyTorch.
    \item Optimiser: Adam with a learning rate of 0.001 and weight decay of $1 \times 10^{-3}$.
    \item Batch size: 128.
\end{itemize}

The following describes the different settings for each training set-up:
In normal \textbf{local training}, the full 50,000 training samples are used for model training, and evaluation is conducted on the entire test set. Since MCUNet-in3 is pre-trained on ImageNet, directly applying it for transfer learning by training only the classifier layer results in poor performance, with test accuracy around 0.40. Therefore, full-model fine-tuning is performed to adapt the model to the CIFAR-10 dataset. After this step, the classifier layer of the final model is replaced and used as the base model for subsequent transfer learning.
In \textbf{CFL training}, we utilised the Flower framework \cite{beutel:hal-03601230}, an open-source platform for building federated learning systems. We tested both IID and Non-IID data distribution modes for device datasets. For Non-IID mode, a fixed random seed was used to maintain consistent data distribution, enabling direct comparison with DFL settings. Model aggregation employed FedAvg, with the evaluation metric being the weighted average of model accuracies, calculated based on the data volume used for training on each device.
\textbf{Training with the bilayer GD-PSGD framework} also tested both IID and Non-IID data distribution modes. This framework introduces additional hyperparameters:
\begin{itemize}
    \item \textbf{DK\_MEANS\_ITERATIONS}: Maximum iterations for DK-means clustering. Approximate convergence is sufficient, so the algorithm is terminated early to save computation. Testing showed that 5 iterations are adequate.
    \item \textbf{INTRA\_GOSSIP\_ROUNDS}: Maximum rounds of intra-cluster communication using the gossip protocol.
    \item \textbf{INTER\_GOSSIP\_ROUNDS}: Maximum rounds of inter-cluster communication using the gossip protocol.
\end{itemize}
For \textbf{both CFL and bilayer GD-PSGD}, the number of local training epochs per device is set to 2. Excessive local training epochs can lead to catastrophic forgetting, where the model forgets previous aggregation results. Testing with 1-6 epochs showed that local models converged fully within 4 epochs, with comparable results between 2 and 4 epochs. Thus, we selected 2 epochs to optimise training speed while maintaining performance.
\section{Results and Discussion}
\begin{figure}
  \centering
    \subfloat[Loss over time]{\includegraphics[width=0.25\textwidth]{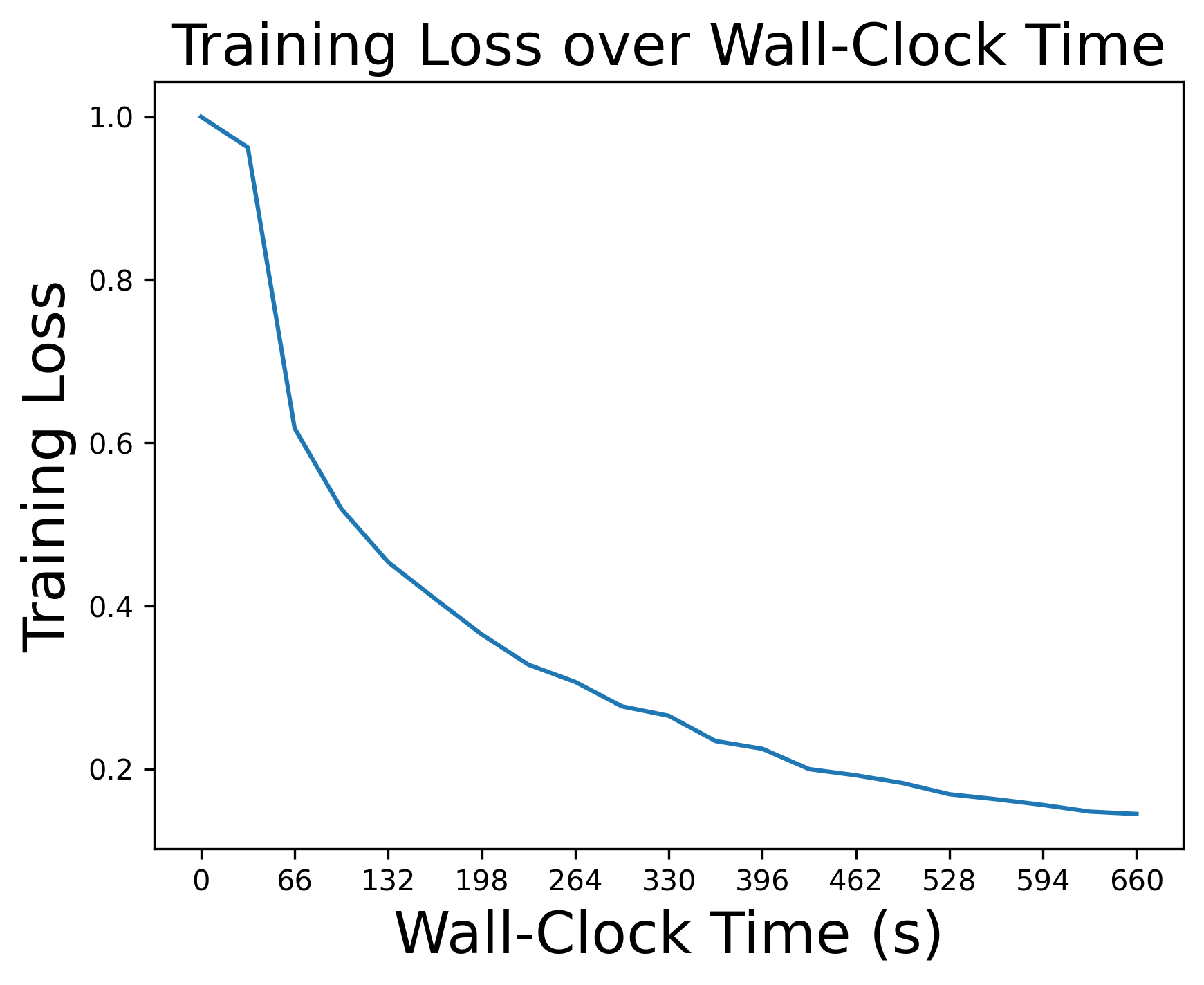}\label{fig:baseline_loss_time}}
    \subfloat[Loss over epochs]{\includegraphics[width=0.25\textwidth]{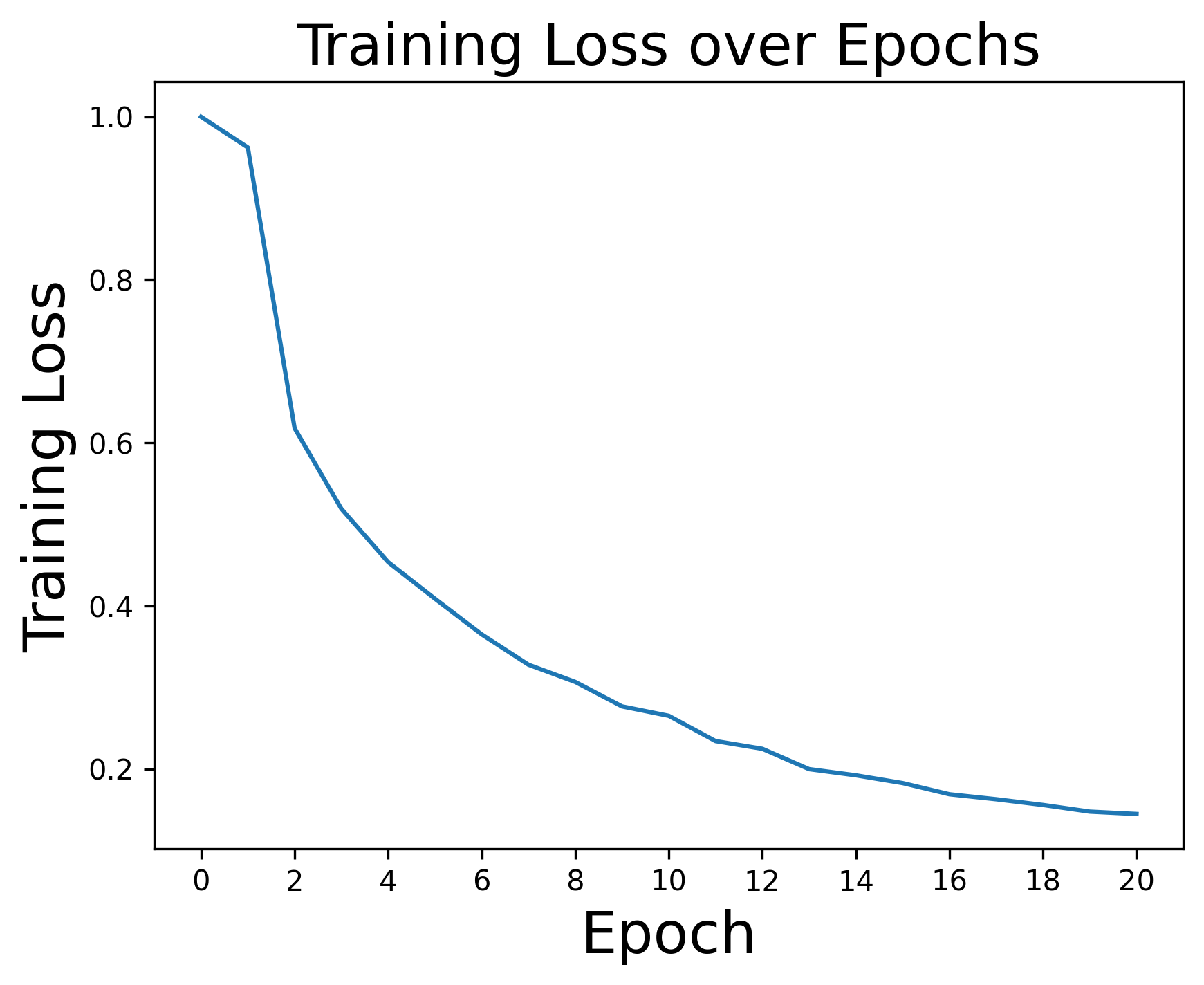}\label{fig:baseline_loss_epoch}} \\
    \subfloat[Accuracy]{\includegraphics[width=0.25\textwidth]{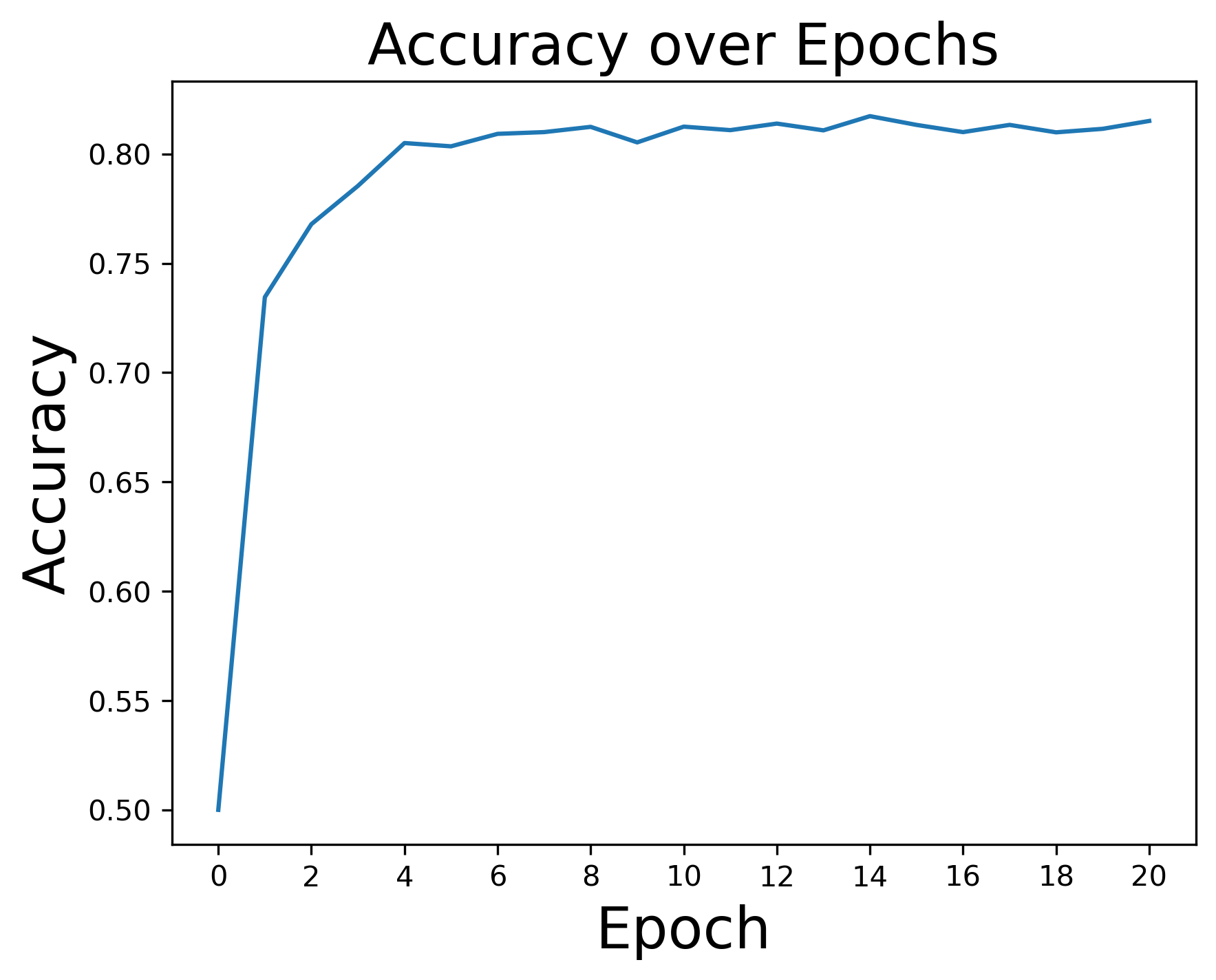}\label{fig:baseline_accuracy}}
    \subfloat[F1 Score]{\includegraphics[width=0.25\textwidth]{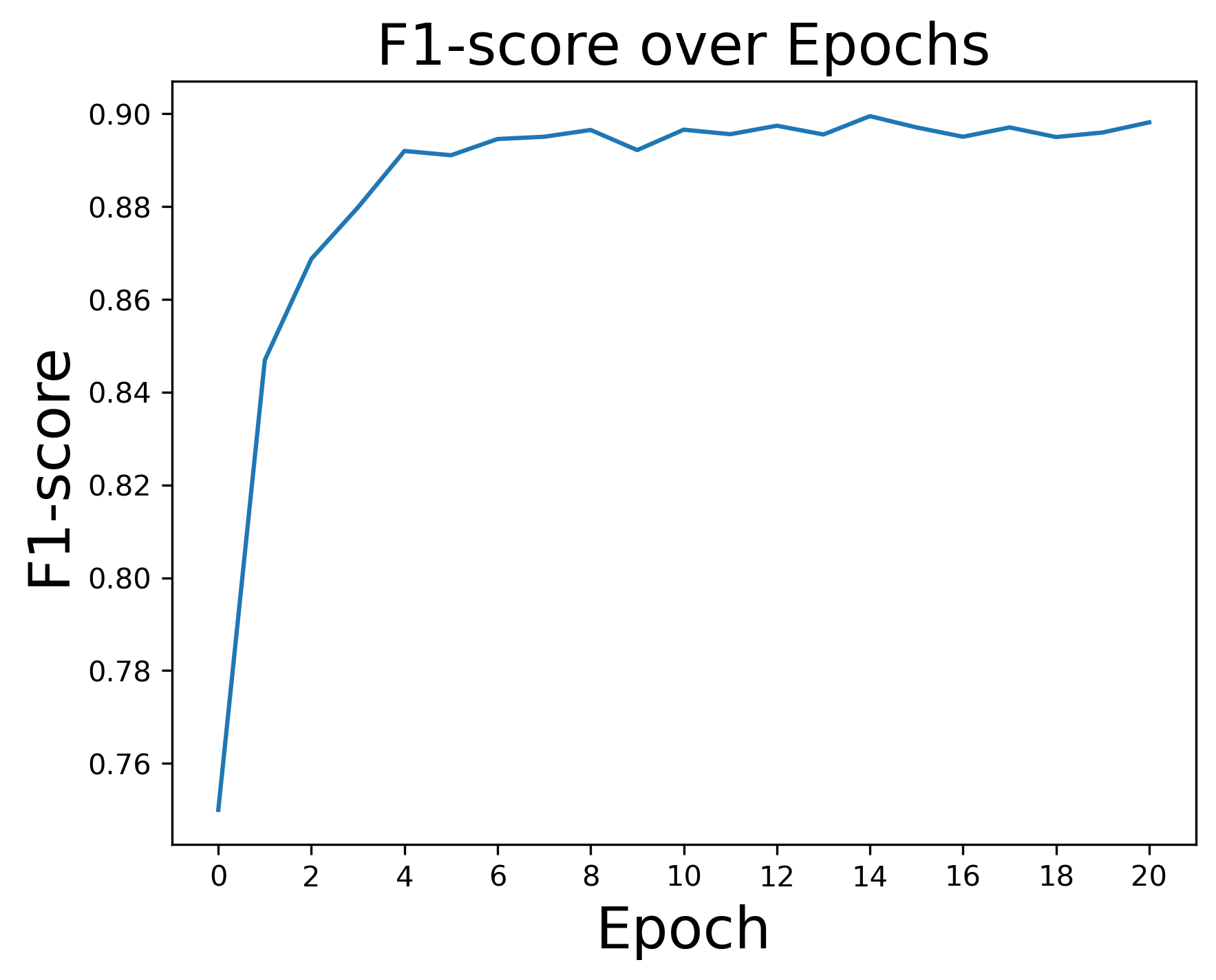}\label{fig:baseline_f1}}
  \caption{Performance of the baseline model on the complete CIFAR-10 dataset.}
  \label{fig:baseline_performance}
\end{figure}

Full-model training was conducted on the complete CIFAR-10 dataset to establish a performance baseline. This baseline serves as an upper bound for comparison but is not directly contrasted with CFL or DFL since it uses the entire dataset for training and does not simulate federated learning conditions. The results are shown in Figure~\ref{fig:baseline_performance}.

The CFL baseline and bilayer GD-PSGD were evaluated under identical conditions:
\begin{itemize}
    \item Optimizer: Adam (learning rate = 0.001, weight decay = 1e-5).
    \item Local training epochs: 2.
    \item Number of devices: 30.
    \item Communication range: 60.
    \item Dataset: IID partition with balanced subsets.
\end{itemize}

\begin{figure}
  \centering
    \subfloat[Accuracy over time]{\includegraphics[width=0.45\textwidth]{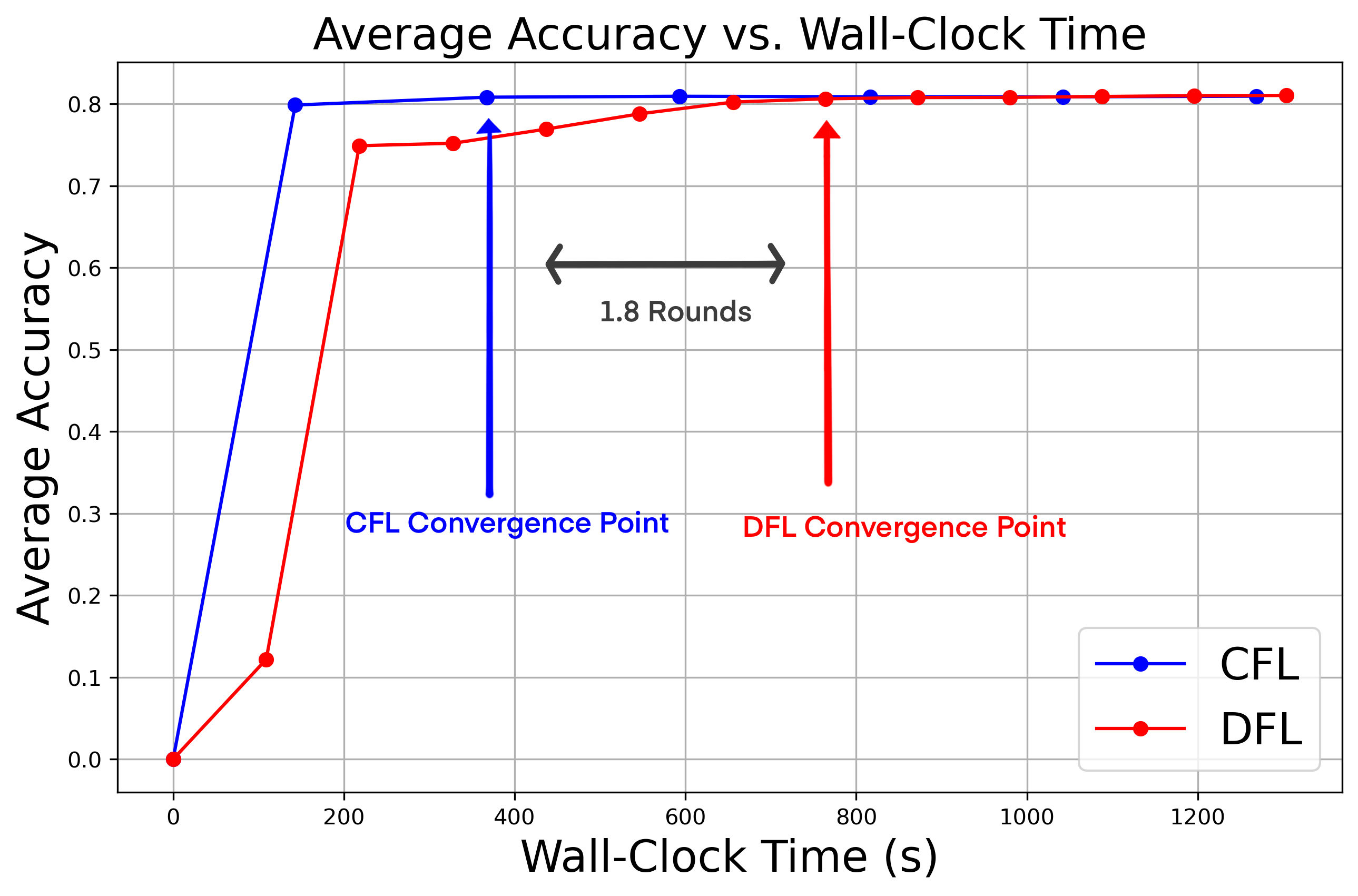}\label{fig:acc_time_30device_balance}} \\
    \subfloat[Accuracy over rounds]{\includegraphics[width=0.45\textwidth]{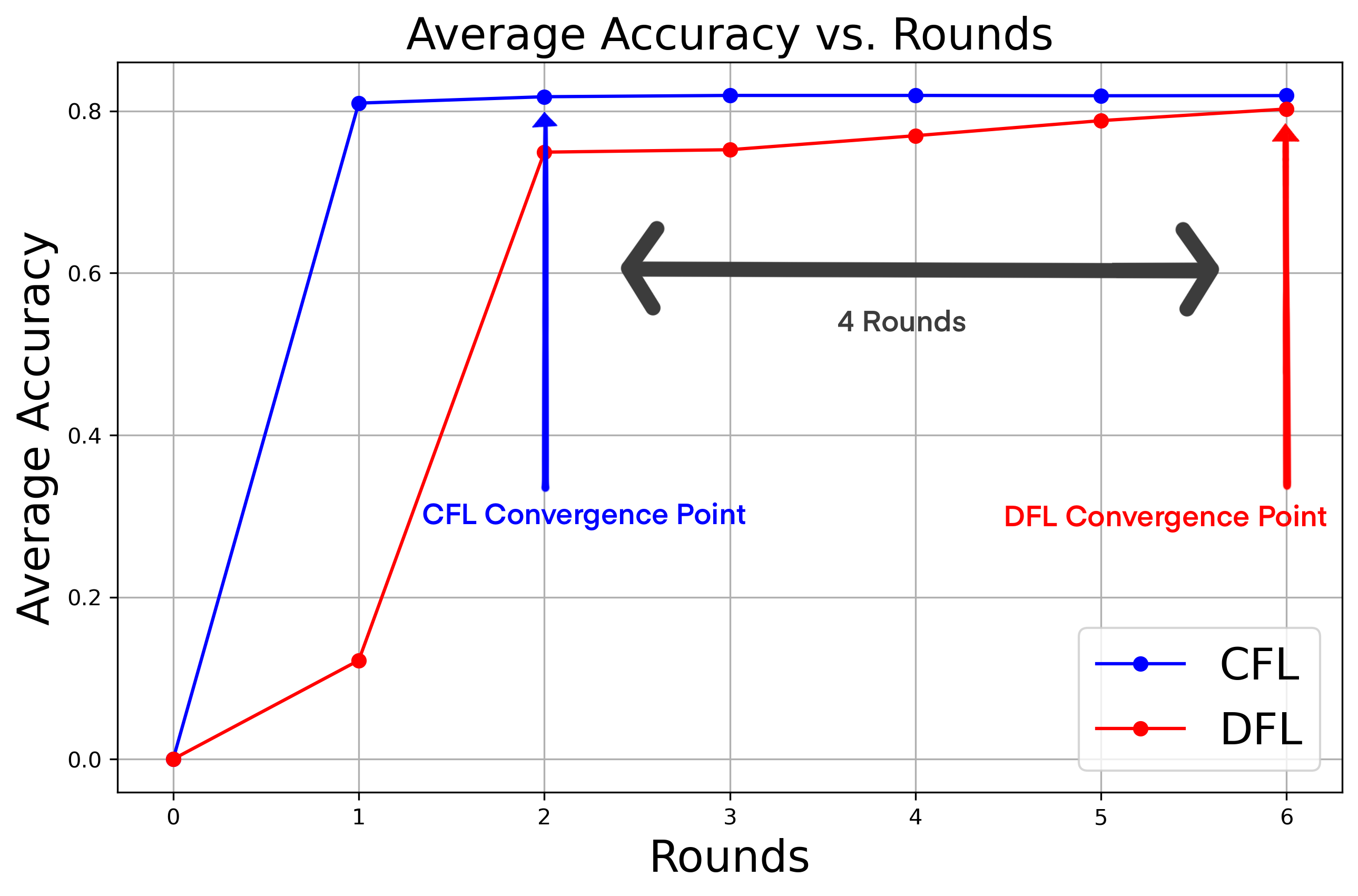}\label{fig:acc_rounds_30device_balance}} \\
    \subfloat[F1 Score over rounds]{\includegraphics[width=0.45\textwidth]{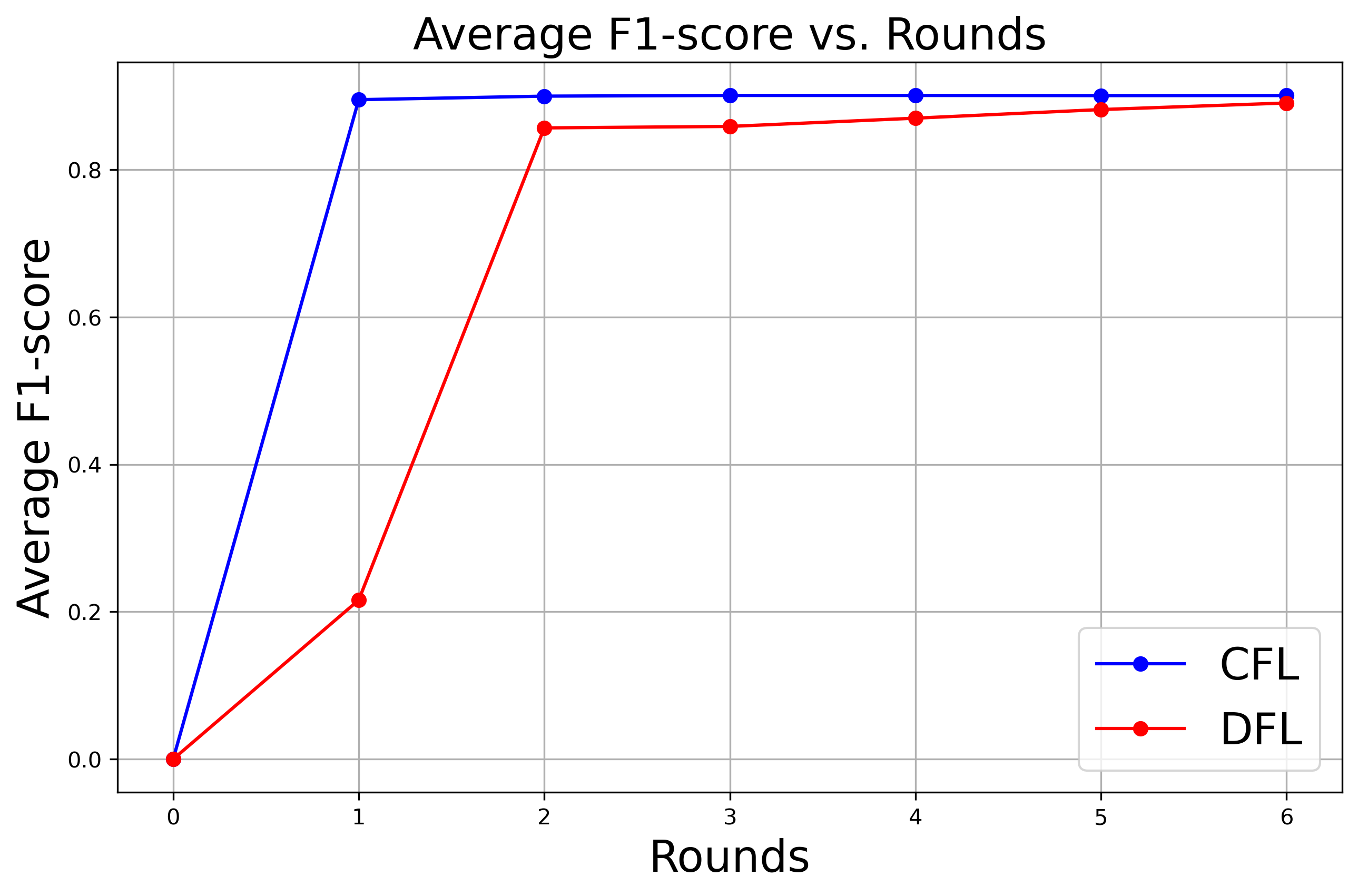}\label{fig:f1_rounds_30device_balance}} \\
    \subfloat[Loss over rounds]{\includegraphics[width=0.45\textwidth]{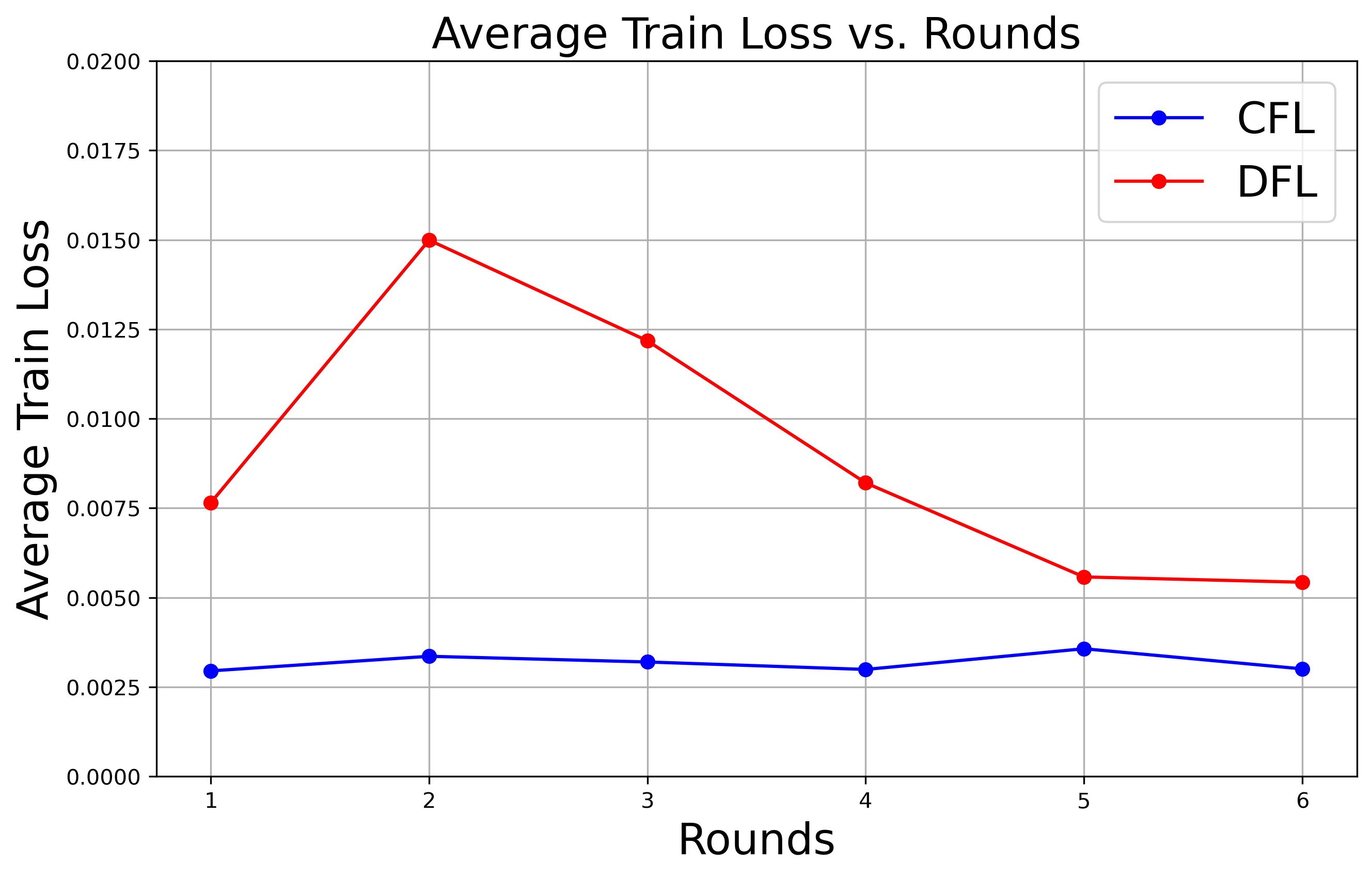}\label{fig:loss_rounds_30device_balance}}
  \caption{Comparison of CFL and bilayer GD-PSGD performance on IID datasets.}
  \label{fig:iid_performance}
\end{figure}

The simulation was conducted sequentially on a single computer to ensure consistency. The results are shown in Figure~\ref{fig:iid_performance}.

\begin{itemize}
    \item CFL converged in a single round, achieving 81.89\% accuracy.
    \item DFL converged in 7 rounds, reaching 81.05\% accuracy, closely matching CFL and the baseline model (81.73\%).
\end{itemize}

Despite requiring more rounds to converge, DFL's per-round computation time was shorter. When measured in wall-clock time, DFL converged in approximately 1.8 CFL-equivalent rounds, as shown in Figure~\ref{fig:acc_time_30device_balance}.
Both CFL and DFL achieved similar final accuracy, demonstrating the robustness of the proposed bilayer GD-PSGD. However, CFL's centralized nature ensures faster convergence, while DFL balances decentralization with competitive performance.
The convergence rate was evaluated using both the number of rounds and wall-clock time. DFL required 4 additional rounds to converge compared to CFL but completed each round faster. When measured by wall-clock time, DFL took slightly longer to converge, demonstrating the trade-off between decentralized computation and speed.

\subsection{Performance on Non-IID Datasets}

\begin{figure}
  \centering
    \subfloat[Accuracy over time]{\includegraphics[width=0.45\textwidth]{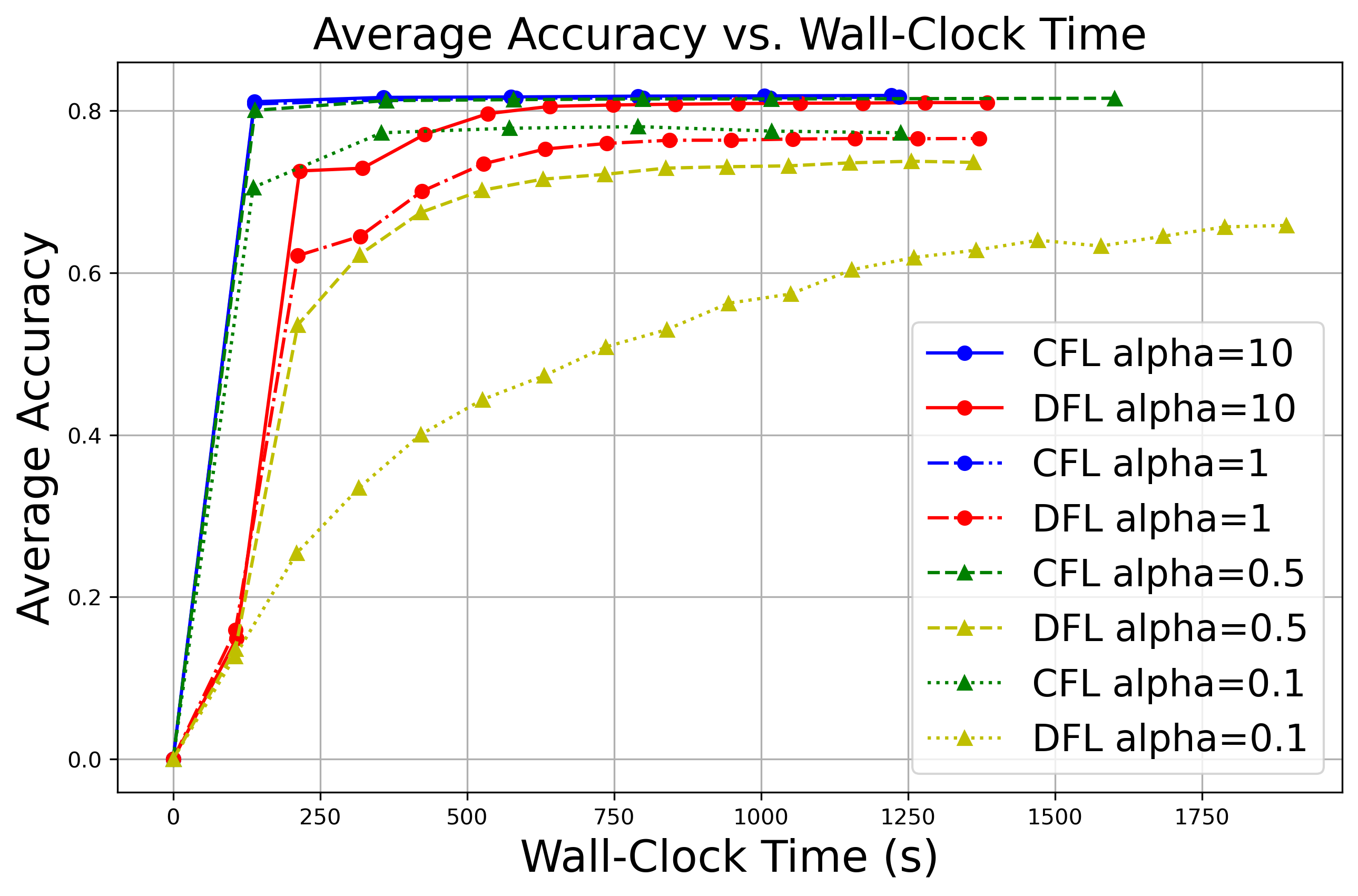}\label{fig:acc_time_30device_alpha}} \\
    \subfloat[Accuracy over rounds]{\includegraphics[width=0.45\textwidth]{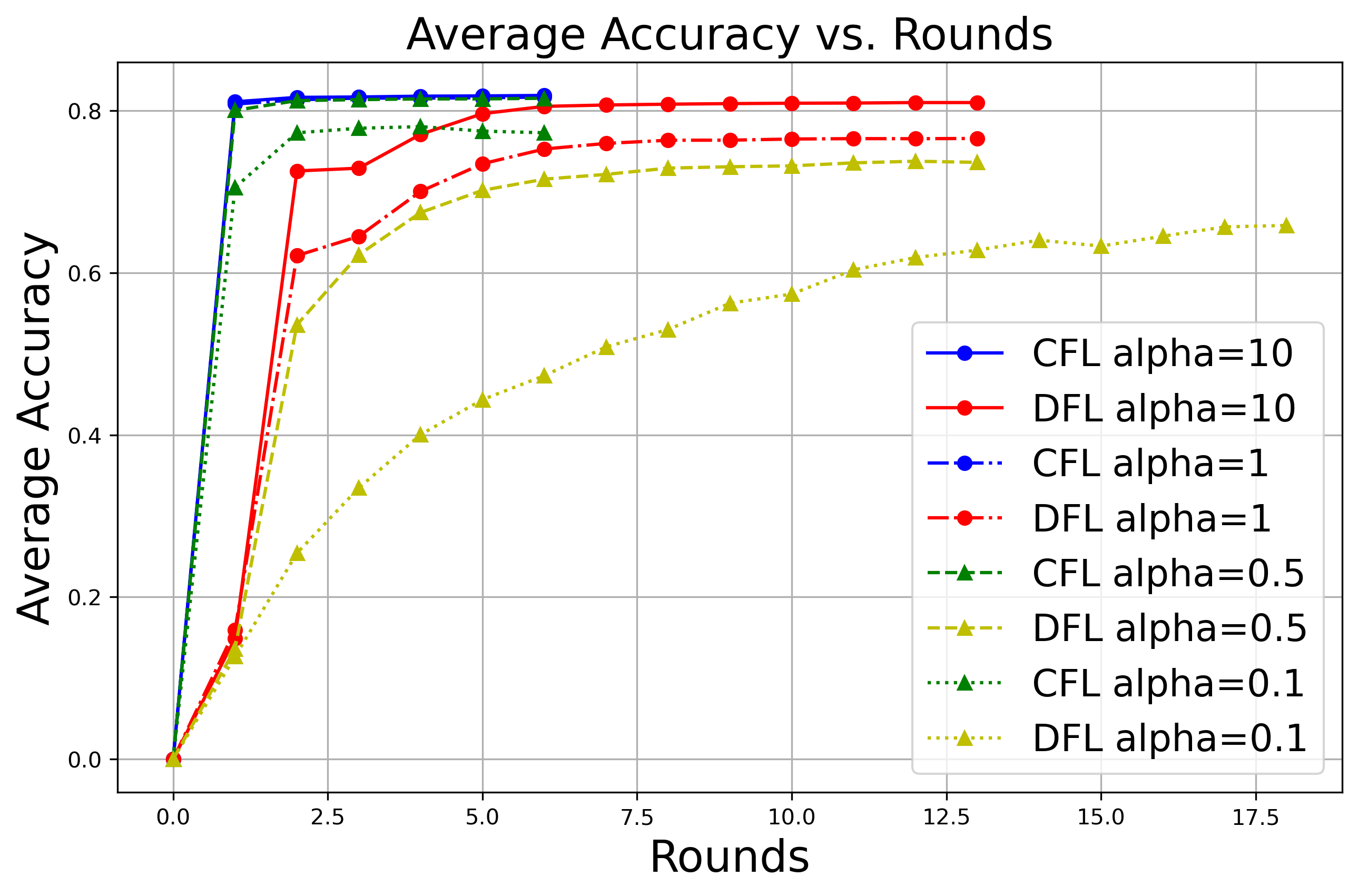}\label{fig:acc_rounds_30device_alpha}} \\
    \subfloat[F1 score over rounds]{\includegraphics[width=0.45\textwidth]{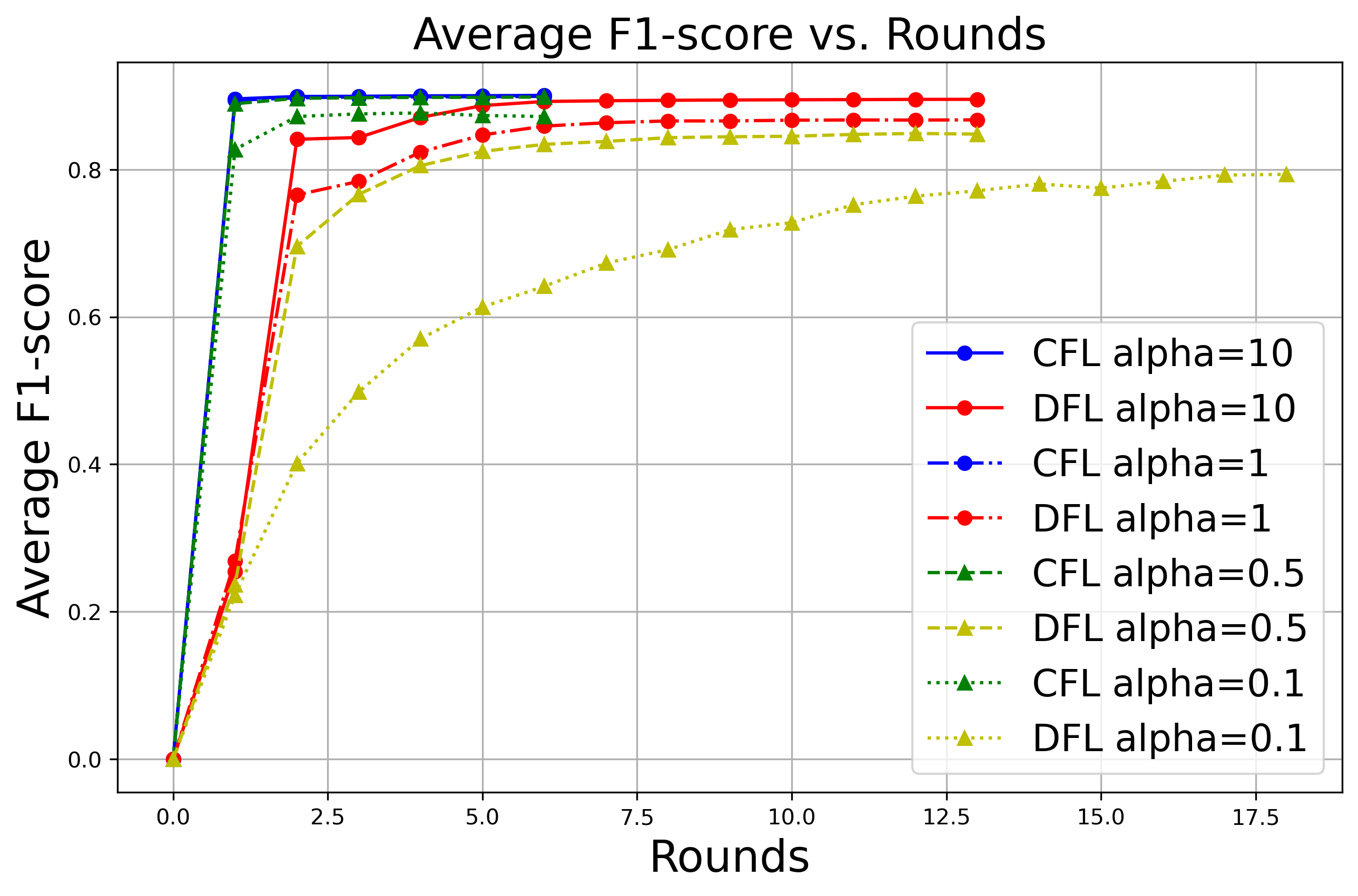}\label{fig:f1_rounds_30device_alpha}} \\
    \subfloat[Loss over rounds]{\includegraphics[width=0.45\textwidth]{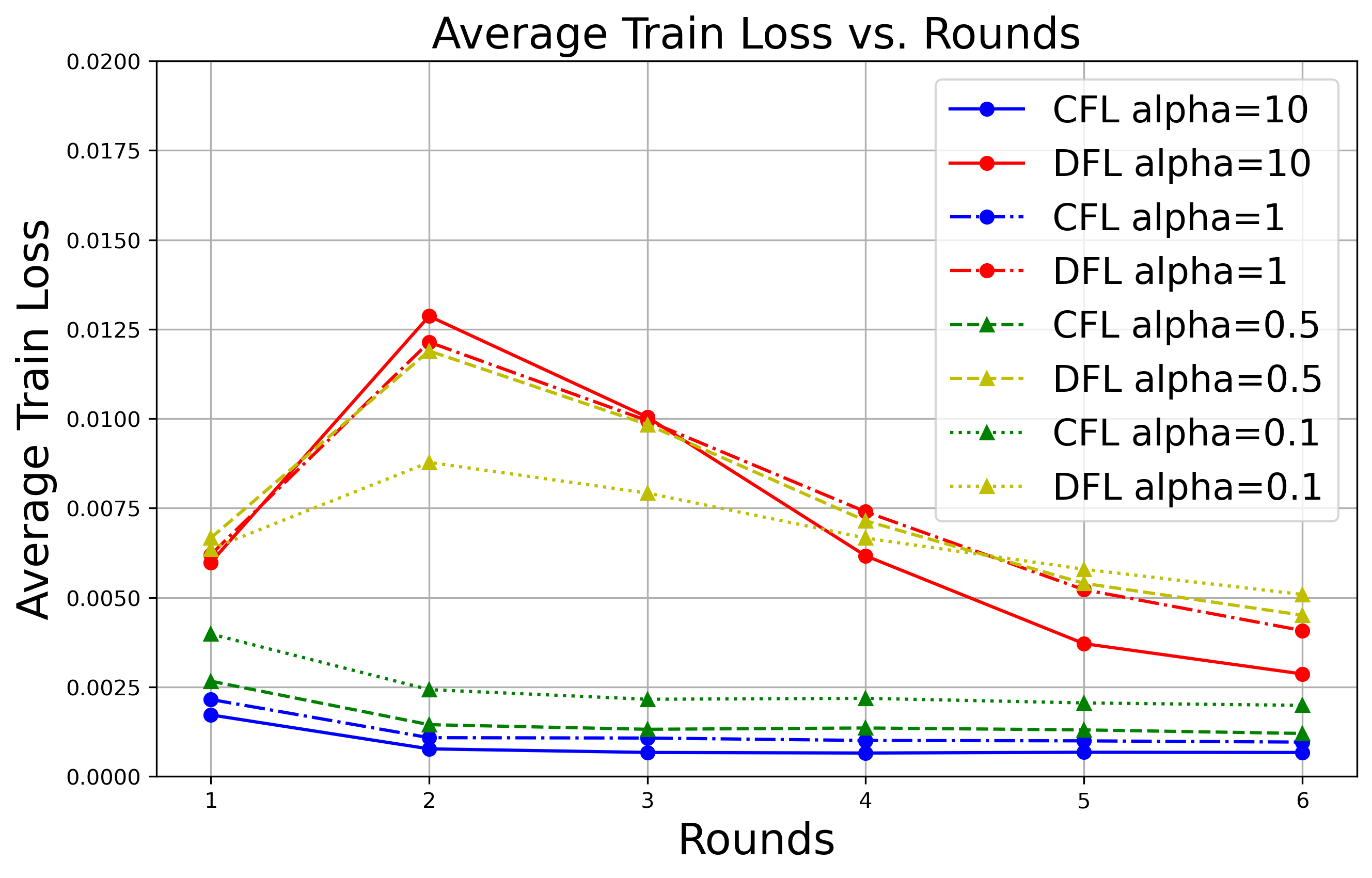}\label{fig:loss_rounds_30device_alpha}}
  \caption{Comparison of CFL and DFL performance at different non-IID levels (\(\alpha\)).}
  \label{fig:noniid_performance}
\end{figure}

Non-IID datasets are a significant challenge in federated learning. This section evaluates whether the proposed DFL method can sustain high performance under non-IID conditions. The parameter \(\alpha\) was used to control the degree of non-IIDness while keeping all other training parameters constant. The results are shown in Figure~\ref{fig:noniid_performance}.

\begin{itemize}
    \item CFL Performance: CFL converged within 1-2 rounds regardless of \(\alpha\), with minimal impact on accuracy until \(\alpha = 0.1\). At \(\alpha = 0.1\), CFL accuracy dropped by 4.4\% to 77.3\%, and convergence slowed by one round.
    \item DFL Performance: DFL showed a gradual decline in convergence rate and final accuracy as \(\alpha\) decreased:
        \begin{itemize}
            \item \(\alpha = 10\): Convergence in 6-7 rounds with 81.0\% accuracy.
            \item \(\alpha = 0.1\): Convergence in 18 rounds with 65.9\% accuracy.
        \end{itemize}
\end{itemize}

\(\alpha\) determines the disparity between local datasets. Lower \(\alpha\) increases disparity, challenging the decentralized nature of DFL:
\begin{itemize}
    \item In CFL, the server aggregates updates from all devices, ensuring comprehensive information propagation.
    \item In DFL, limited communication range restricts information exchange, particularly for isolated or skewed devices.
\end{itemize}

Zhao et al. \cite{DBLP} attribute accuracy drops to weight divergence, which also affects DFL architectures. While CFL mitigates this issue through global aggregation, DFL faces difficulties in highly non-IID settings.
\subsection{Impact of Number of Devices on DFL Performance}

\begin{figure}
    \centering
    \includegraphics[width=0.45\textwidth]{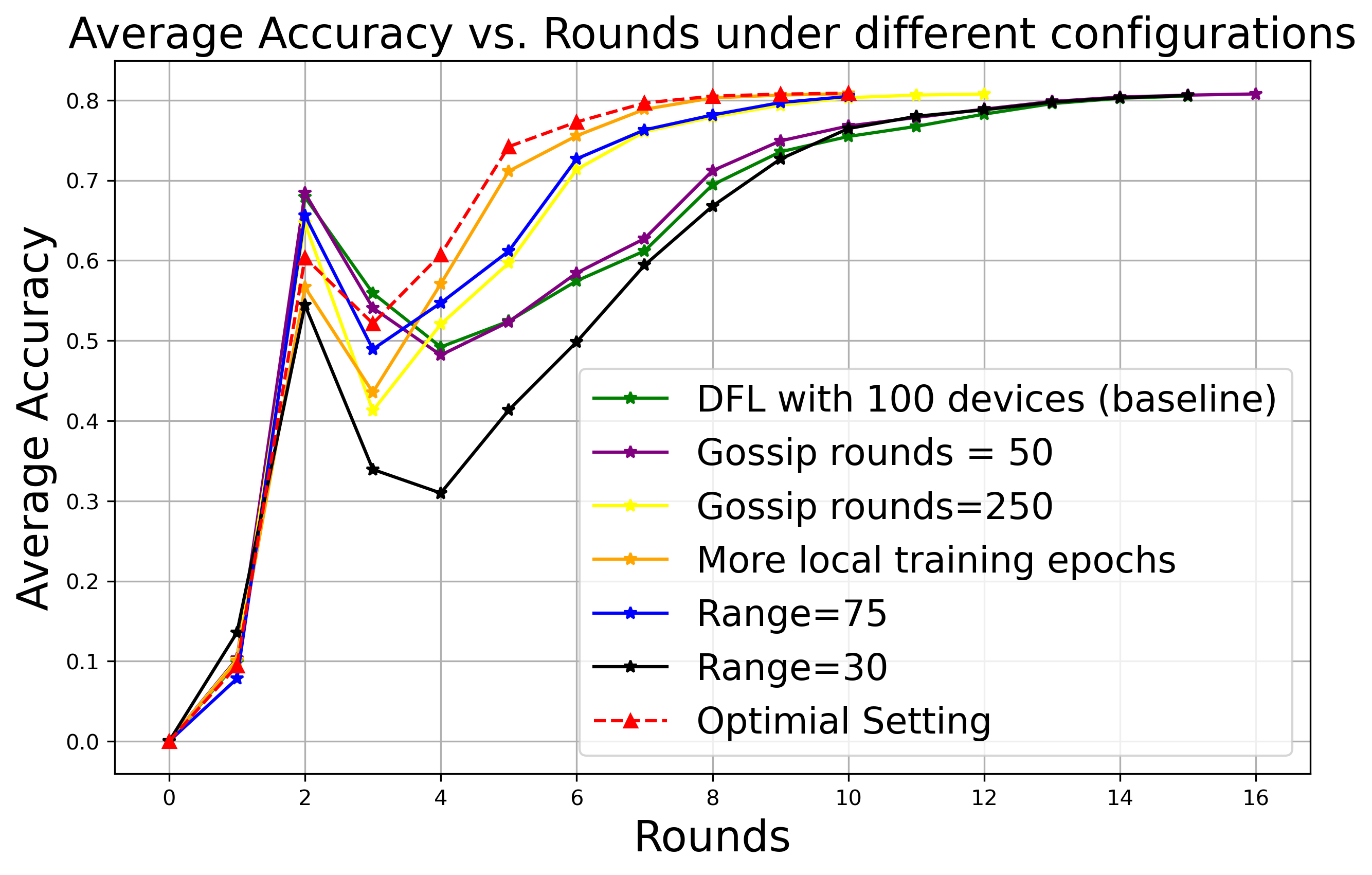}
    \caption{Effect of communication range, local training epochs, and gossip rounds on DFL convergence with 100 devices.}
    \label{fig:acc_rounds_100device_all}
\end{figure}

The number of devices significantly influences DFL performance. Increasing the number of devices within the same simulation area leads to:
\begin{itemize}
    \item Smaller local datasets per device, increasing the complexity of training.
    \item A denser network topology, enhancing information dissemination.
\end{itemize}
\begin{figure}
  \centering
    \subfloat[Accuracy over rounds]{\includegraphics[width=0.45\textwidth]{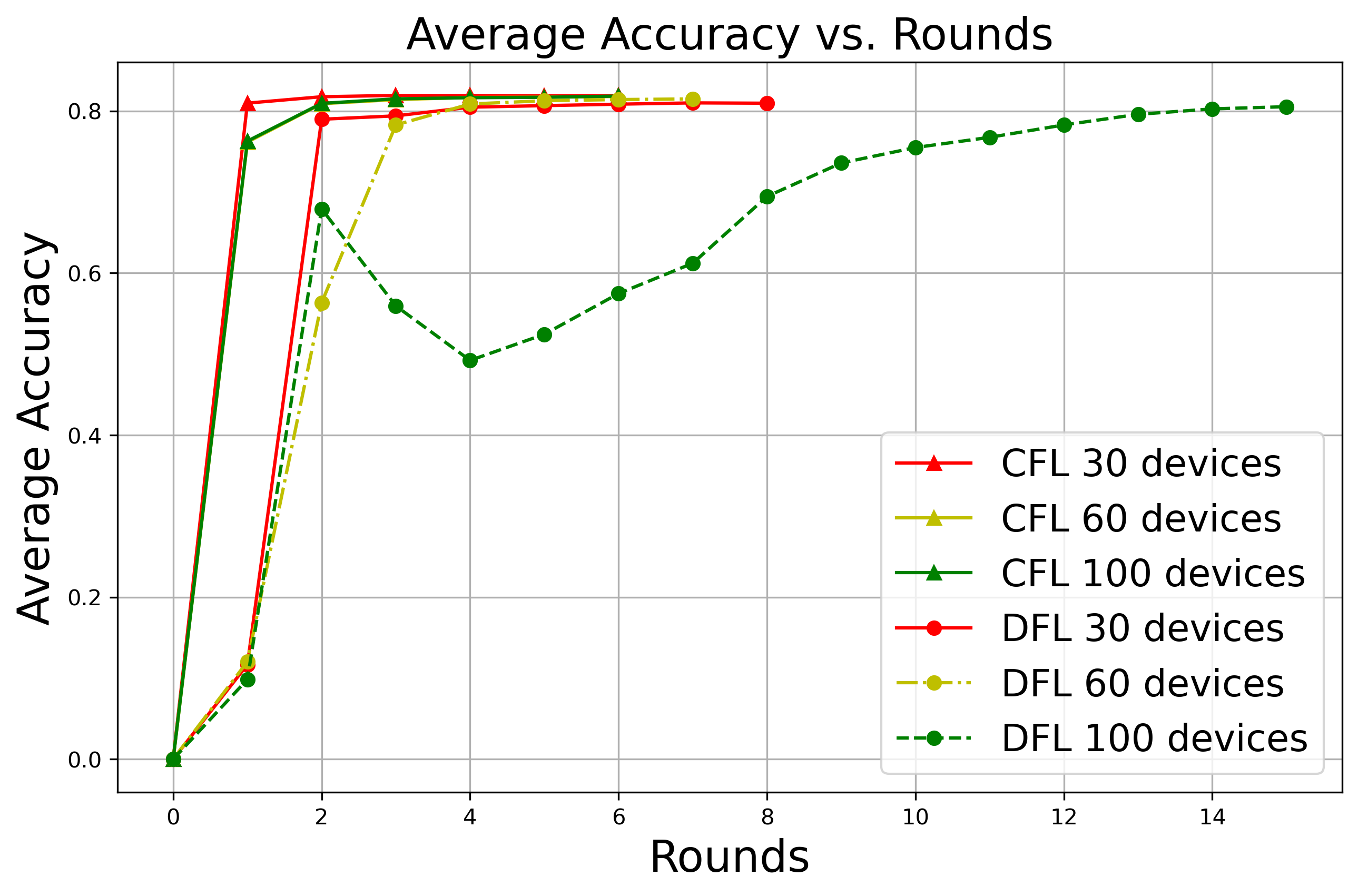}\label{fig:acc_rounds_num_device}} \\
    \subfloat[Loss over rounds]{\includegraphics[width=0.45\textwidth]{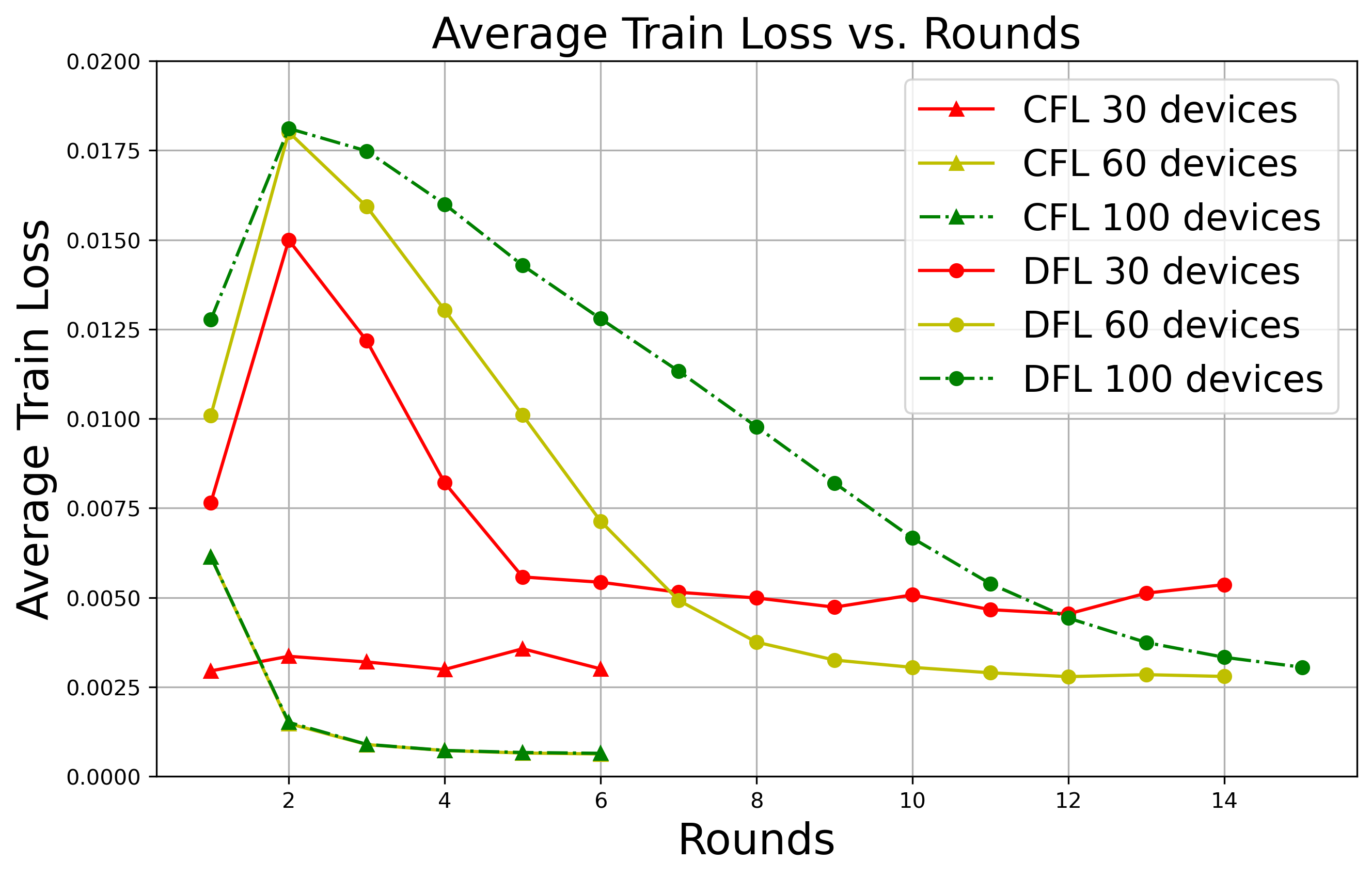}\label{fig:loss_rounds_30device_num_device}}
  \caption{Comparison of CFL and DFL performance with varying device counts.}
  \label{fig:num_devices_performance}
\end{figure}
To evaluate the effect of device count, we conducted experiments and compared convergence rates by the number of rounds instead of wall-clock time (Figure~\ref{fig:num_devices_performance}), with the following observations:
\begin{itemize}
    \item CFL Performance: Convergence rates were consistent across device counts, and accuracy was unaffected.
    \item DFL Performance: Increasing device counts slowed convergence. With 100 devices, accuracy fluctuated before stabilizing at 80.5\% in the 14th round.
    \item Larger device counts exacerbate information lag and overfitting in DFL due to limited local data and communication delays.
    \item Despite slower convergence, DFL achieved similar final accuracy across device counts, demonstrating its robustness.
    \item Dense networks unexpectedly slowed convergence, likely due to differences in gossip protocol implementations compared to prior studies \cite{pmlr-v97-koloskova19a}.
    \item At 100 devices, observed accuracy fluctuations were likely caused by local overfitting and delayed information dissemination.
\end{itemize}
\vspace{0.5em}
Further experiments with controlled variable adjustments confirmed that increasing local training epochs, communication range, and gossip rounds improved convergence but did not completely eliminate fluctuations (Figure~\ref{fig:acc_rounds_100device_all}).

\subsection{Impact of Communication Range on DFL Performance}

The communication range significantly affects the network topology in DFL. A shorter communication range:
\begin{itemize}
    \item Reduces the number of neighbouring devices within the range.
    \item Lowers the degree of freedom for each node in the network.
\end{itemize}

The degree of freedom affects the spectral properties of the gossip or Laplace matrix. As the spectral gap increases (i.e., the difference between the largest and second-largest eigenvalues), well-connected graphs exhibit faster information exchange. Conversely, sparsely connected nodes exchange information more slowly, delaying convergence.

\begin{figure}
  \centering
    \subfloat[Accuracy over ranges]{\includegraphics[width=0.45\textwidth]{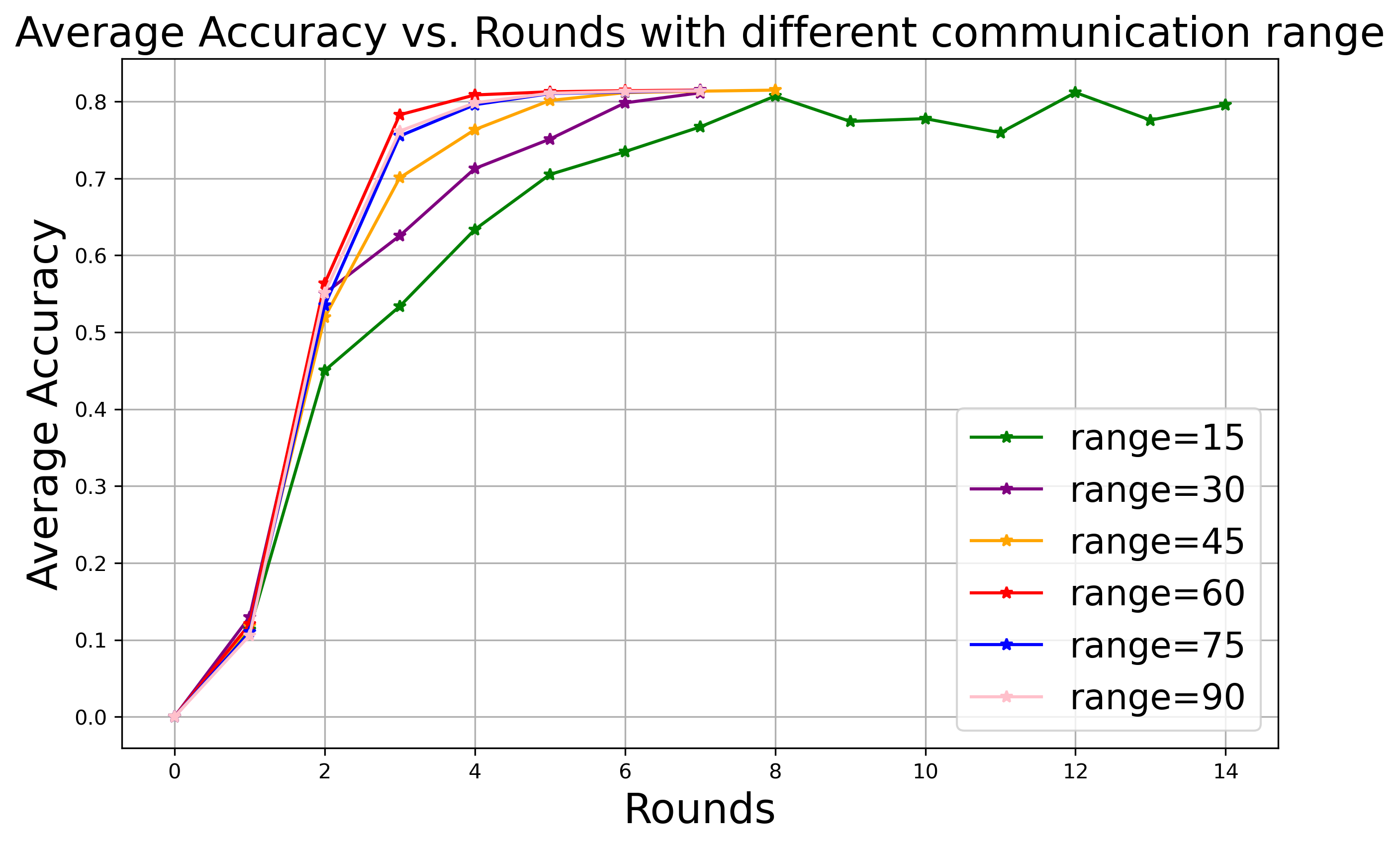}\label{fig:acc_rounds_range}} \\
    \subfloat[Loss over ranges]{\includegraphics[width=0.45\textwidth]{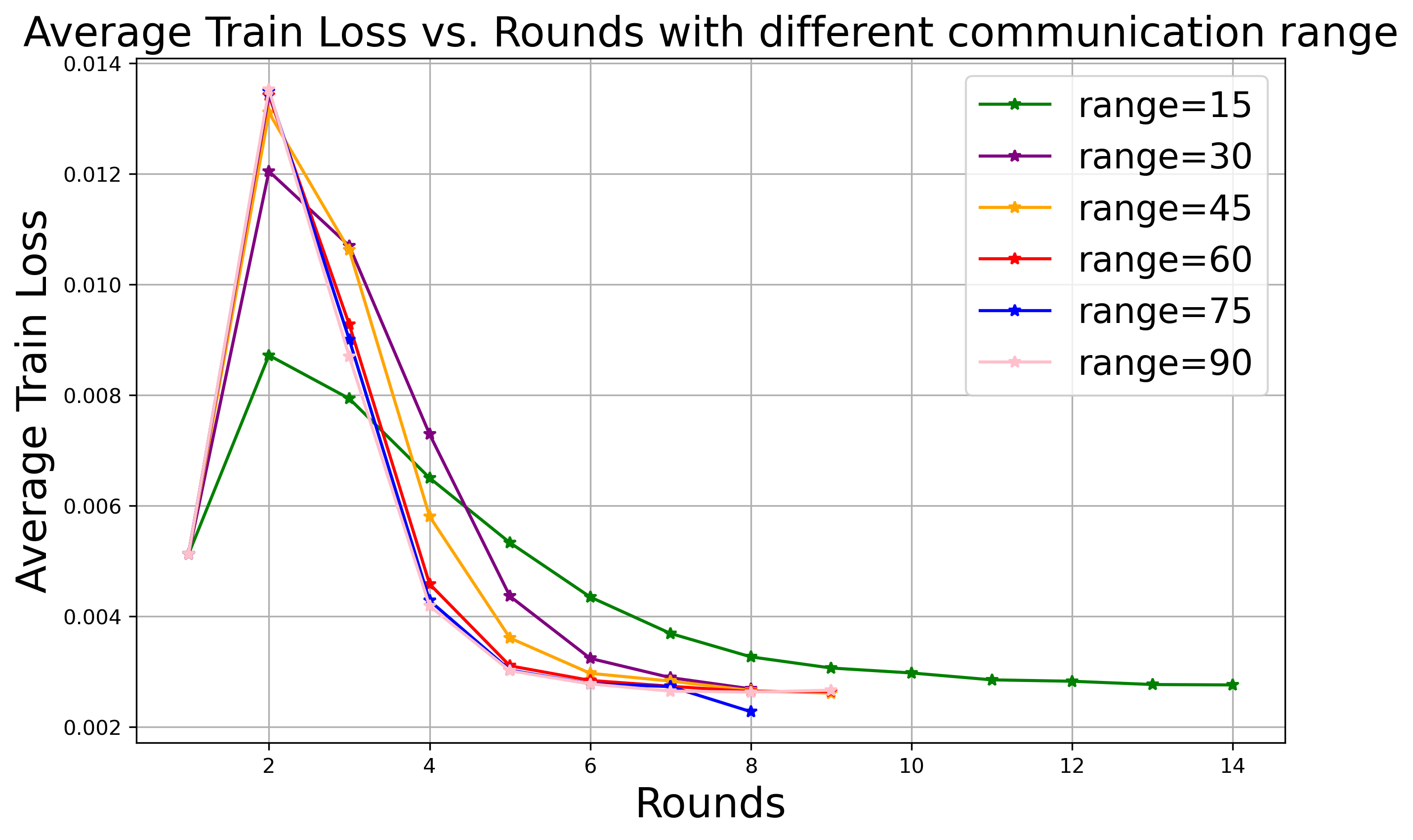}\label{fig:loss_rounds_range}}
  \caption{Comparison of CFL and DFL performance at varying communication ranges.}
  \label{fig:range_performance}
\end{figure}

In our experiments, we fixed the number of devices at 60, maintaining a moderate density within the simulation area. Results are shown in Figure~\ref{fig:range_performance} with the following observations:.

\begin{itemize}
    \item Short Ranges (15):The model reached 80\% accuracy in 8 rounds but exhibited unstable performance with fluctuations over the next 3 rounds.
    \item Moderate Ranges (30-45):Convergence occurred within 6-7 rounds, with stable performance after convergence.
    \item Long Ranges (60+):Convergence occurred within 4-5 rounds. Increasing the range beyond 60 had no further impact on convergence.
\end{itemize}

\begin{figure}
  \centering
    \subfloat[Accuracy on non-IID data]{\includegraphics[width=0.45\textwidth]{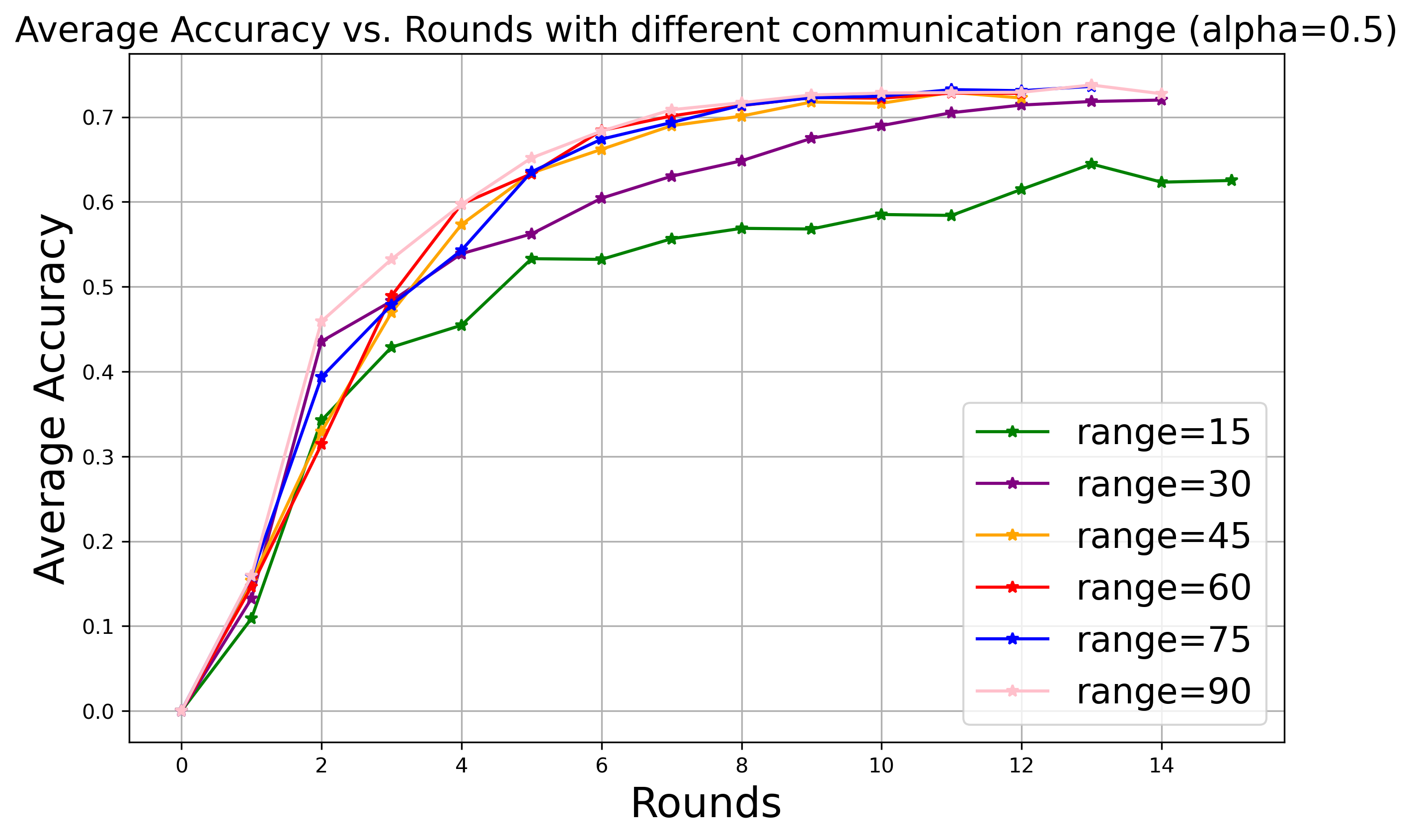}\label{fig:acc_rounds_alpha_range}} \\
    \subfloat[Loss on non-IID data]{\includegraphics[width=0.45\textwidth]{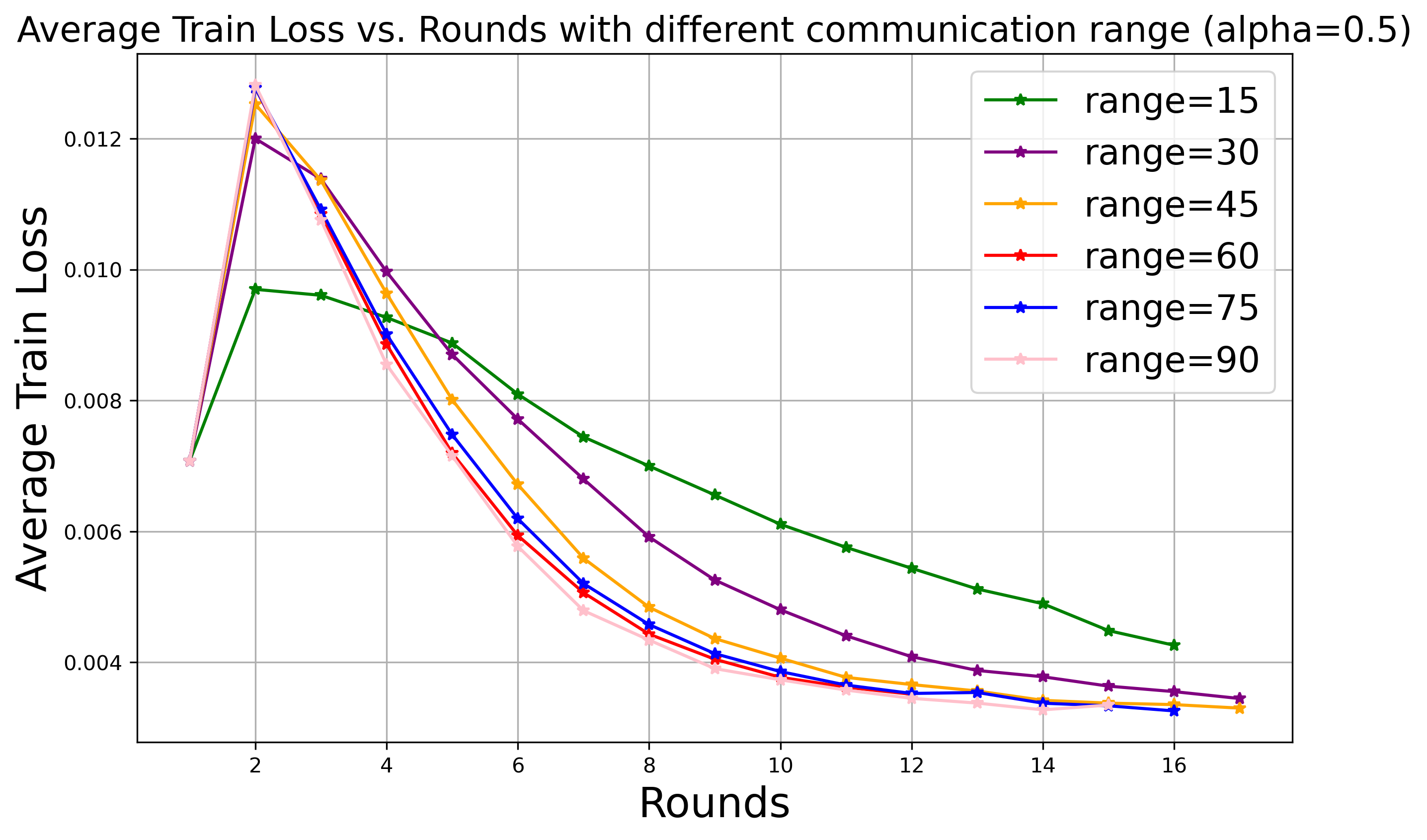}\label{fig:loss_rounds_05_range}}
  \caption{Impact of communication ranges on DFL performance with non-IID datasets.}
\end{figure}

Once the communication range exceeds a certain threshold, further increases do not significantly impact convergence rate or accuracy. This threshold balances information dissemination and network density.
In non-IID settings, longer communication ranges improved convergence and accuracy. As shown in Figure~\ref{fig:acc_rounds_alpha_range}, when the range exceeded 45, performance reached its upper limit. Increasing the range further had minimal impact.

\subsection{Impact of Geographic Location on DFL Performance }

We also explored whether clustering devices based on their geographic location is the optimal criterion. Studies by Duan et al. and Zhao et al. propose using EMD (Earth Mover's Distance) to gauge dataset similarity and aggregate weights of devices with similar data distributions \cite{10203996, DBLP}. This approach has shown improved global model performance, particularly for non-IID settings. In our experiments, we used EMD values as the clustering criterion for DK-means to assess its impact.

\begin{figure}
    \centering
    \includegraphics[width=0.45\textwidth]{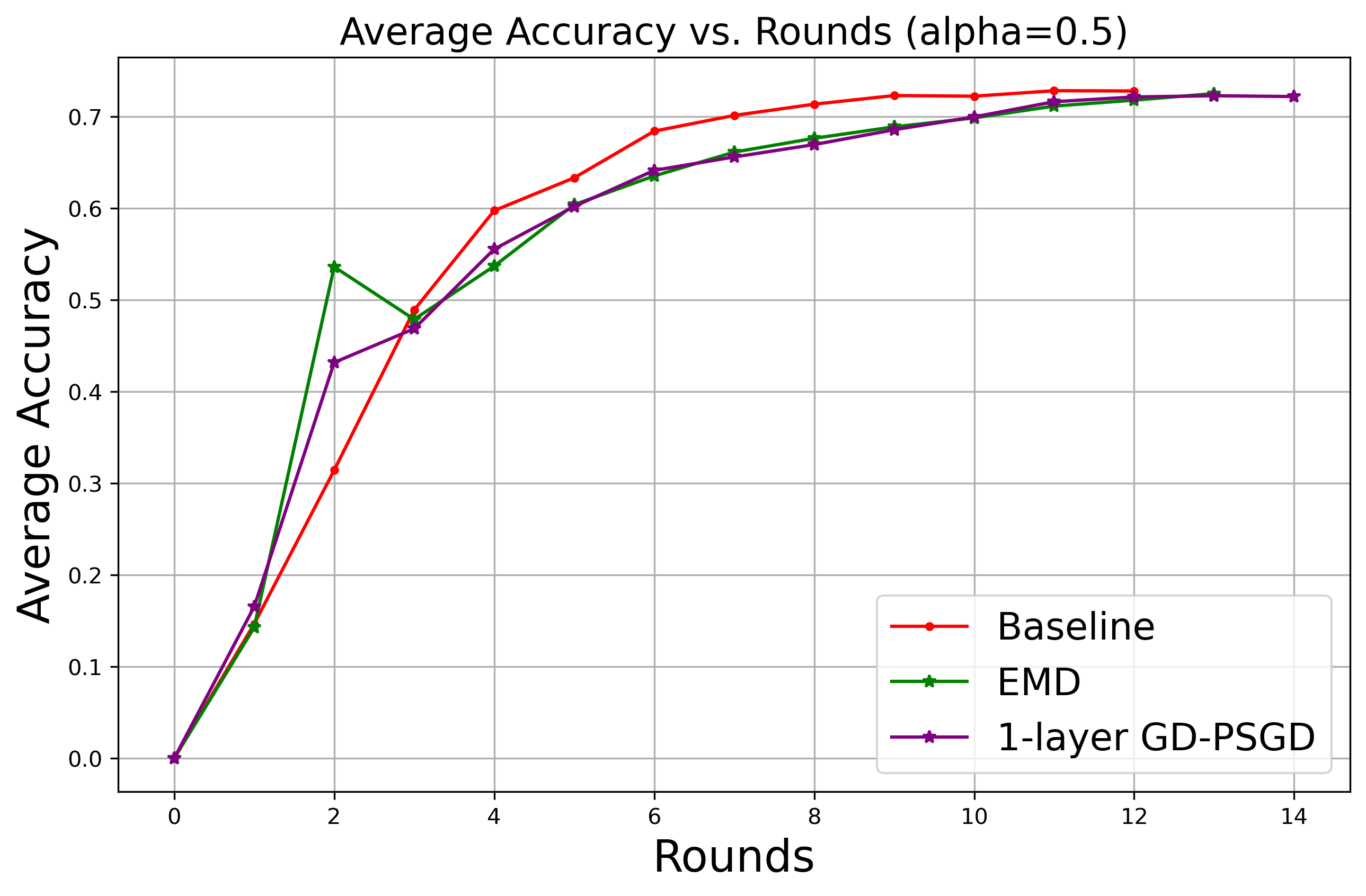}
    \caption{Comparison of clustering criteria: geographic location vs. EMD values. Baseline settings: number of devices=60, \(\alpha=0.5\), communication range=60.}
    \label{fig:additional_exp}
\end{figure}

Figure \ref{fig:additional_exp} shows that while the final accuracy of the EMD-based variant remains comparable, it does not converge as quickly as the bilayer GD-PSGD. These findings suggest:
\begin{itemize}
    \item The bilayer network topology significantly accelerates convergence in GD-PSGD.
    \item Using EMD values for clustering does not improve the convergence rate in the simulation scenarios considered in this study.
\end{itemize}

\begin{figure}[h!]
    \centering
    \includegraphics[width=0.45\textwidth]{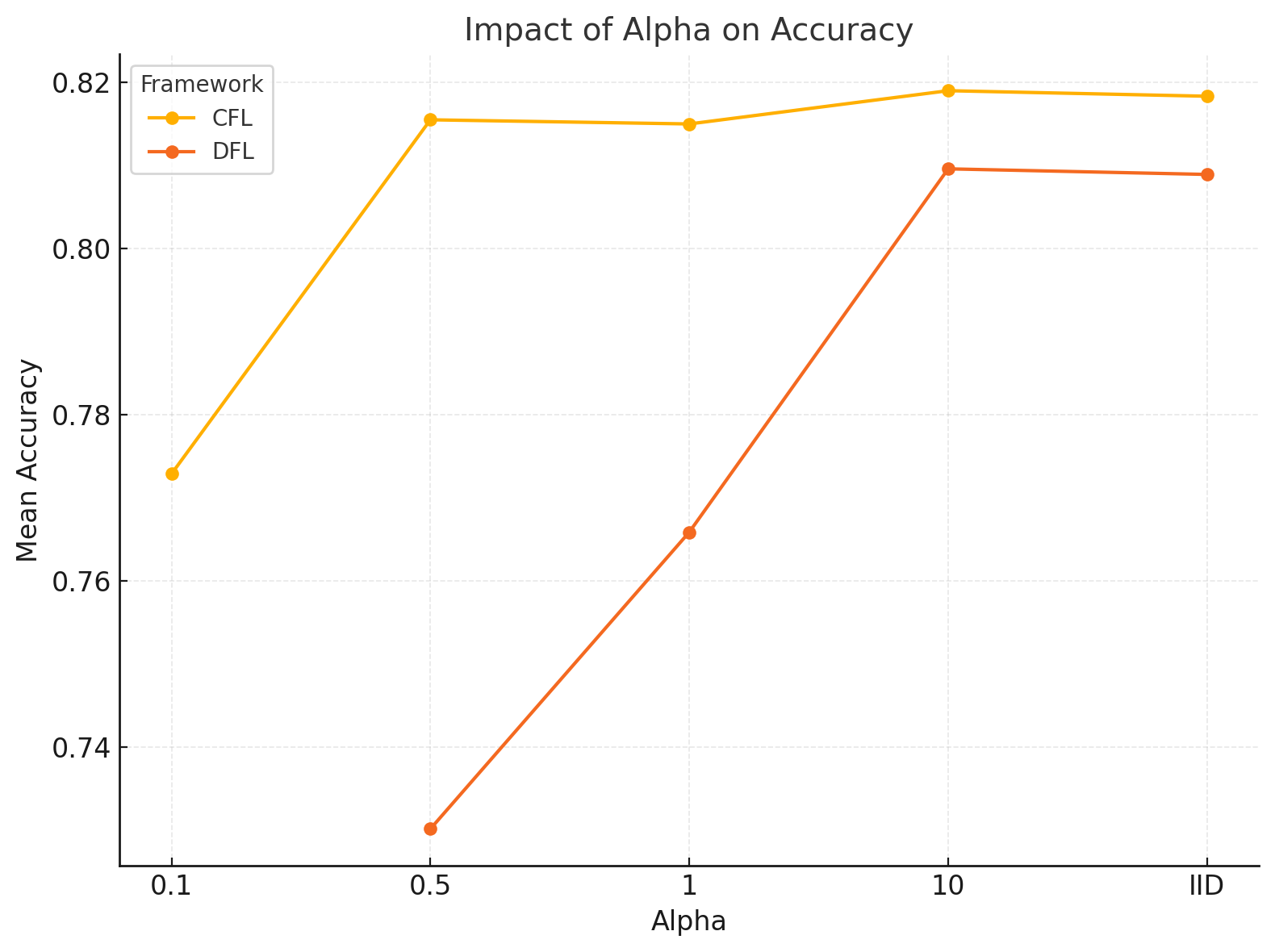}
    \caption{Impact of Alpha on Accuracy for CFL and DFL Frameworks.}
    \label{fig:impact_alpha_accuracy}
\end{figure}

\section{Limitations and opportunities}

Decentralised federated learning (DFL) faces notable challenges including the absence of standardised benchmarks and evaluation tasks. The diverse design possibilities in DFL—such as varying network topologies, communication protocols, and deployment conditions—make comparisons difficult. Differences in static versus dynamic networks and edge device capabilities further complicate consistent evaluation. As Martínez et al. \cite{Mart_nez_Beltr_n_2023} highlight, these disparities hinder fair comparisons and the establishment of clear state-of-the-art benchmarks.

The second limitation is the lack of real-world testing instances on edge devices. Microcontrollers, the typical hardware for such deployments, often lack the resources for full-model training. While Nvidia Jetson devices support full-model training, their cost and size make them unsuitable for widespread deployment. MIT's TinyEngine enables training on microcontrollers with less than 256KB memory but supports only heavily quantised MCUNet models and STM32CubeIDE, making deployment cumbersome \cite{256kb}.

Addressing these limitations is essential for advancing DFL and enabling its broader adoption.
The rapid advancement of DFL and edge device technology suggests promising directions for future work. Real-world deployment of DFL architectures and on-device training on edge devices should be explored. With improved hardware supporting full model training, future studies can bypass transfer learning for end-to-end model training.

Moreover, integrating bilayer, multilayer, or hierarchical aggregation mechanisms into DFL offers significant potential. These methods, proven effective in CFL architectures, could further enhance DFL convergence rates. Future research should investigate diverse hierarchical topologies and criteria, such as geolocation, Earth Mover’s Distance (EMD), or combinations of both, to optimise network architectures \cite{9699080} \cite{abs-1910-01991}.

\section{Conclusion}

This study introduces bilayer GD-PSGD (Gossip Decentralized Parallel SGD), a distributed federated learning framework designed for edge devices in remote areas. The framework addresses challenges such as limited computational resources, constrained communication ranges, lack of network coverage, and dynamic device networks. Using the DK-means algorithm, devices are clustered based on geographic locations, followed by gossip protocol for communication and cumulative FedAvg for model aggregation across intra-cluster and inter-cluster layers.

Experiments using the MCUNet-in3 model and CIFAR-10 dataset compared bilayer GD-PSGD with the traditional CFL architecture. Key findings include:

\begin{itemize}
    \item On IID datasets, bilayer GD-PSGD matches CFL’s final performance (81\% accuracy) with only 1.8 additional rounds required for convergence.
    \item On Non-IID datasets, bilayer GD-PSGD retains high performance, with less than 8\% accuracy loss for \(\alpha = 0.5\). Severe imbalance (\(\alpha = 0.1\)) causes a performance drop of about 16\%.
    \item Communication range minimally impacts final performance, even on Non-IID datasets. Only a range of 15 produces an accuracy degradation of less than 8\%.
    \item Increasing the number of devices slightly slows convergence but does not affect final performance. For over 100 devices, fine-tuning hyperparameters reduces convergence rounds to 8.
\end{itemize}

These results highlight bilayer GD-PSGD’s ability to closely match CFL’s performance while demonstrating robustness across varying datasets, communication ranges, and device numbers.


\bibliographystyle{IEEEtran}
\bibliography{sample}

\appendix 
\textbf{Properties of Relevant Gossip Protocol}:
The  paper also solicits a background on gossip model. Bracha protocol  is one of the  prominent example of Gossip protocol in distributed systems.
The well known Bracha protocol results in O(n2) message complexity. Conventionally,the analysis on such protocol deploys the possibility to reduce it to $O(\sqrt{n})$.
without depending on quorum but on group construction of size (o(n)
and also scrutinize  the possibility to get a honest majority to agree on the message of the sender.
To explain the main idea of the  specific  protocol, we first recall the crucial steps that allow the agreement on a message m sent by a transmitter p in Bracha’s RBC described  for the definition of reliable broadcast and Bracha’s protocol\cite{bra}. Bracha’s RBC assume a fixed number of nodes n, where fraction of f are byzantine. In each step nodes send a single message. To progress, each node must receive enough messages from the other nodes con- firming that they have seen the message from the sender, it is possible that in the first step (echo) the nodes see conflicting messages, the second step (ready) allows to confirm the reception of at least (n f )/2 that have completed the first step (echo) or that are already in the last step (accept) to decide on the value of the message sent by the sender, the second step is crucial it allows to decide on the value that will eventually be accepted in the accept step. Finally, if the node receives enough ready (the nodes that have done the second step) it accepts the message.

\textbf{Scaleability of Bracha model}:
Broadcast proposed by Bracha scale poorly, since it depends strongly on quorum, which grows linearly with the number of nodes. To overcome quorum dependency, \cite{Guerraoui19}, replace quorum by an extensive use of samples, via a random oracle placed locally in each processor and pick a random node identities with uniform probability. The size of the samples is chosen so as to to decrease the possibility of violating the broadcast primitive. The authors consider a modular approach that depends mainly on two components: the first one is Probabilistic broadcast called Murmur, whose objective is to gossip messages (each node communicates with	(log n) nodes and receives (log n) messages) and to ensure the notion of totality, roughly speaking it means that the majority of nodes receive messages from the sender. This part does not ensure consistency (agree on a single message) since if the sender is byzantine the nodes can observe different messages. To overcome this the authors propose Sieve, the second component, where nodes use extensively the oracle that proposes identity of nodes with replacement called echo sample, with whom to send the message and wait to observe enough echo of the same message from their sample to deliver it. This guarantees consistency (agreeing on the same message) in a group of nodes that share the same echo. However, if the sender is byzantine, it is possible with a non-negligible probability that different messages are accepted in different echo groups. The main algorithm of the probabilistic reliable broadcast abstraction called Contagion, is inspired from Barcha, realizes simultaneously the totality and the consistency with a high probability, where roughly speaking Sieve and Murmur are used in the first two steps of Bracha broadcast. To perform the final step, a correct process again uses the oracle to choose a sample by repeatedly call the oracle to send a ready messages to and wait to collect a specific size of ready messages to accept Contagion.



\sloppy
\textbf{Lemma:} Let $n$ denote the number of participants in a distributed system, and $L$ represent the latency of a single message exchange. Consider two protocols:

\begin{enumerate}
    \item Protocol A with communication complexity $O(n)$.
    \item Protocol B with communication complexity $O(\sqrt{n})$.
\end{enumerate}

Then, we assert the following:

\begin{enumerate}
    \item \textbf{Lower Latency:} Protocol B achieves lower latency compared to Protocol A as $n$ increases. Proof: $T_A = n \times L$, $T_B = \sqrt{n} \times L$, and $T_B$ grows slower than $T_A$ due to the square root relationship.
    
    \item \textbf{Improved Throughput:} Protocol B achieves higher throughput as $n$ increases because throughput is inversely proportional to latency.
    
    \item \textbf{Better Resource Utilization:} Protocol B utilizes network bandwidth and processing resources more efficiently as $n$ increases due to fewer message exchanges.
    
    \item \textbf{Scalability:} Protocol B exhibits better scalability as $n$ increases because its communication complexity grows slower, making it more scalable in large-scale distributed systems.
\end{enumerate}

Therefore, achieving $O(\sqrt{n})$ communication complexity results in lower latency, improved throughput, better resource utilization, and enhanced scalability in distributed systems.

\section{Introduction}

\textbf{Lemma: Quorum Formation and Dependency Tracking in Distributed Systems}

In a distributed system, quorum formation and dependency tracking play essential roles in achieving consensus, fault tolerance, and reliability. The formula $|Q| = g(N, f, |D|)$ represents the relationship between quorum formation ($Q$), system parameters including the total number of nodes ($N$), maximum number of faulty nodes ($f$), and the number of dependencies ($|D|$), where $g(N, f, |D|)$ is a function that determines the size and composition of quorums.

\begin{enumerate}
    \item \textbf{Minimum Quorum Size}: The function $g(N, f, |D|)$ ensures that each quorum contains a minimum number of nodes required for agreement, considering fault tolerance requirements. It guarantees that quorums are sufficiently large to tolerate up to $f$ faulty nodes while maintaining agreement.
    
    \item \textbf{Dependency-Aware Quorum Formation}: The function $g(N, f, |D|)$ incorporates dependency tracking information to adjust quorum compositions dynamically. It ensures that dependencies between messages or events are satisfied by including the necessary nodes in each quorum, thereby guaranteeing agreement on dependent events.
    
    \item \textbf{Fault Tolerance Mechanisms}: The function $g(N, f, |D|)$ accounts for fault tolerance mechanisms to handle failures effectively. It dynamically adjusts quorum sizes or compositions in response to node failures or network partitions, ensuring system availability and consistency.
    
    \item \textbf{Quorum Intersection}: The function $g(N, f, |D|)$ ensures that quorums intersect sufficiently to prevent system partitioning. It calculates quorum compositions in a way that guarantees quorum intersection, maintaining consistency and liveness guarantees.
    
    \item \textbf{Recovery Strategies}: The function $g(N, f, |D|)$ defines recovery strategies to restore the system to a consistent state after failures or disruptions. It reconfigures quorums, resynchronizes nodes, or reestablishes communication channels to recover from failures effectively, ensuring system reliability.
\end{enumerate}

By satisfying these properties, the formula $|Q| = g(N, f, |D|)$ provides a comprehensive framework for designing quorum formation mechanisms that ensure correctness, fault tolerance, and reliability in distributed systems.

\vspace{2ex}
\noindent \textbf{Mathematical Models and Formal Representations for Optimization Techniques}

\subsection{\textbf{Hierarchical Structuring}}
\textbf{Mathematical Model:} Let \( C \) represent the set of clusters, and \( E_c \) represent the set of edges connecting clusters. The goal is to minimize the communication and connectivity costs:
\[
\text{Minimize } \text{Comm}(C) + \text{Conn}(E_c)
\]
\textbf{Formal Representation:}
\[
f_{\text{hier}}(C, E_c) = \min_{C, E_c} \text{Comm}(C) + \text{Conn}(E_c)
\]

\subsection{\textbf{Overlay Networks}}
\textbf{Mathematical Model:} Let \( G = (V, E) \) represent the physical network, and \( G' = (V', E') \) represent the overlay network. The objective is to minimize the overlay distance and efficiency:
\[
\text{Minimize } \text{Dist}(E') + \text{Efficiency}(G', E')
\]
\textbf{Formal Representation:}
\[
f_{\text{overlay}}(G, G') = \min_{G', E'} \text{Dist}(E') + \text{Efficiency}(G', E')
\]

\subsection{\textbf{Partial Replication}}
\textbf{Mathematical Model:} Let \( D \) represent the set of data items, and \( R \) represent the set of nodes for replication. The aim is to minimize the replication cost:
\[
\text{Minimize } \text{ReplicationCost}(R)
\]
\textbf{Formal Representation:}
\[
f_{\text{replication}}(D, R) = \min_{R} \text{ReplicationCost}(R)
\]

\subsection{\textbf{Optimized Message Dissemination}}
\textbf{Mathematical Model:} Let \( M \) represent the set of messages, and \( N \) represent the set of nodes. The objective is to minimize the message overhead of the dissemination algorithm:
\[
\text{Minimize } \text{MessageOverhead}(\text{Algorithm})
\]
\textbf{Formal Representation:}
\[
f_{\text{dissemination}}(M, N) = \min_{\text{Algorithm}} \text{MessageOverhead}(\text{Algorithm})
\]

\subsection{\textbf{Asynchronous Communication}}
\textbf{Mathematical Model:} Let \( \Delta t \) represent the time delay, and \( \epsilon \) represent the synchronization error. The goal is to minimize synchronization overhead:
\[
\text{Minimize } \text{SyncOverhead}(\Delta t, \epsilon)
\]
\textbf{Formal Representation:}
\[
f_{\text{async}}(\Delta t, \epsilon) = \min_{\Delta t, \epsilon} \text{SyncOverhead}(\Delta t, \epsilon)
\]

\subsection{\textbf{Caching and Prefetching}}
\textbf{Mathematical Model:} Let \( D \) represent the set of data items, and \( N \) represent the set of nodes. The aim is to minimize resource utilization:
\[
\text{Minimize } \text{ResourceUtilization}(D, N)
\]
\textbf{Formal Representation:}
\[
f_{\text{cache}}(D, N) = \min_{D, N} \text{ResourceUtilization}(D, N)
\]

\subsection{\textbf{Dynamic Quorum Adaptation}}
\textbf{Mathematical Model:} Let \( Q \) represent the set of quorums, and \( F \) represent the set of failed nodes. The goal is to minimize the quorum adjustment cost:
\[
\text{Minimize } \text{QuorumAdjustmentCost}(Q, F)
\]
\textbf{Formal Representation:}
\[
f_{\text{quorum}}(Q, F) = \min_{Q} \text{QuorumAdjustmentCost}(Q, F)
\]

\end{document}